\documentclass[a4paper,fleqn]{cas-dc}

\usepackage[numbers]{natbib}
\usepackage{enumitem} 
\usepackage{capt-of}
\def\tsc#1{\csdef{#1}{\textsc{\lowercase{#1}}\xspace}}
\tsc{WGM}
\tsc{QE}

\begin{document}
\let\WriteBookmarks\relax
\def\floatpagepagefraction{1}
\def\textpagefraction{.001}

\shorttitle{ObjMST: Object-focused Multimodal Style Transfer}    

\shortauthors{Chanda et al.}  

\title [mode = title]{ObjMST: Object-focused Multimodal Style Transfer}  
% Object focused? Object based?

\tnotemark[1] 

\tnotetext[1]{} 

      % style=chinese,
      % auid=000,
      % bioid=1,
      % prefix=Sir,
      % orcid=0000-0000-0000-0000,
      % facebook=<facebook id>,
      % twitter=<twitter id>,
      % linkedin=<linkedin id>,
      % gplus=<gplus id>]

\author[1]{Chanda Grover Kamra}%[<options>]
\cormark[1]
\fnmark[1]
\ead{chanda.grover_phd19@ashoka.edu.in}
\ead[url]{https://chandagrover.github.io/}
\credit{Conceptualization, Methodology, Draft writing, Reviewing and Editing}
\affiliation[1]{organization={Ashoka University},
            addressline={Rajiv Gandhi Education City}, 
            city={Sonipat},
%          citysep={}, % Uncomment if no comma needed between city and postcode
            postcode={131029}, 
            state={Haryana},
            country={India}}
% \cortext[1]{Corresponding author}
\author[2]{Indra Deep Mastan}%[]
% Footnote of the second author
% Email id of the second author
\ead{indra.cse@itbhu.ac.in}
\credit{Reviewing, Investigation, Supervision, Methodology}
\affiliation[2]{organization={Indian Institute of Technology, BHU},
            addressline={}, 
            city={Varanasi},
%          citysep={}, % Uncomment if no comma needed between city and postcode
            postcode={221005}, 
            state={Uttar Pradesh},
            country={India}}
\author[1]{Debayan Gupta}%[]
% Footnote of the second author
% \fnmark[3]
% Email id of the second author
\ead{debayan.gupta@ashoka.edu.in}
\credit{Supervision, Investigation, Reviewing}
\cortext[1]{Corresponding author}

\fntext[1]{}

\begin{abstract}
We propose ObjMST, an object-focused multimodal style transfer framework that provides separate style supervision for salient objects and surrounding elements while addressing alignment issues in multimodal representation learning. Existing image-text multimodal style transfer methods face the following challenges: (1) generating non-aligned and inconsistent multimodal style representations; and (2) content mismatch, where identical style patterns are applied to both salient objects and their surrounding elements.
Our approach mitigates these issues by: (1) introducing a Style-Specific Masked Directional CLIP Loss, which ensures consistent and aligned style representations for both salient objects and their surroundings; and (2) incorporating a salient-to-key mapping mechanism for stylizing salient objects, followed by image harmonization to seamlessly blend the stylized objects with their environment. 
We validate the effectiveness of ObjMST through experiments, using both quantitative metrics and qualitative visual evaluations of the stylized outputs. Our code is available at: https://github.com/chandagrover/ObjMST.
\end{abstract}

\begin{highlights}
\item We propose a multimodality-guided ObjMST framework for image style transfer.
\item We generate style encodings using cross-modal StyleGAN inversion via CLIP embeddings.
\item Issue: Alignment issues; sol: masked directional CLIP loss.
\item Issue: Content mismatch; sol: Salient-To-Key attention mechanism. 
\item Quantitative and qualitative results show that ObjMST outperforms relevant baselines.
\end{highlights}
\maketitle
% Main text
\vspace{-0.3cm}
\section{Introduction}
\label{sec:intro}
% {\color{red}[cite]}
Classical Image-guided Image Style Transfer (IIST) methods primarily rely on a style image to supervise the stylization process~\cite{DBLPBatziouIPVK23, liu2021adaattn, kwon2022clipstyler, kamra2023sem}. In contrast, multimodal learning~\cite{LIU202317} integrates additional modalities, such as textual descriptions  \cite{radford2021learning,patashnik2021styleclip} or audio cues \cite{DBLPKangKS10} to guide or augment the stylization \cite{kwon2022clipstyler,kamra2023sem}. There also exist uni-modal (image or text)~\cite{liu2021adaattn, kwon2022clipstyler, kamra2023sem} and multimodal (image and text)~\cite{Wang2024WACV} guided methods for IST.
 
IIST methods that rely solely on image data are limited in practical use when style images are unavailable~\cite{kwon2022clipstyler} or when unique styles are lacking~\cite{Wang2024WACV}. Consequently, there has been a shift towards Text-Guided Image Style Transfer (TIST) methods~\cite{kwon2022clipstyler, kamra2023sem, patashnik2021styleclip}. However, TIST approaches often introduce textual artifacts~\cite{kwon2022clipstyler} due to challenges in achieving effective style alignment and content preservation~\cite{kwon2022clipstyler, kamra2023sem}.

In a recent study, Wang et al.~\cite{Wang2024WACV} proposed a novel cross-modal GAN inversion-based style generation mechanism. This mechanism is guided by both style text and style image, ensuring consistency across modalities.

\begin{figure*}
    \centering
    \begin{minipage}{0.112\linewidth}
    \centering
    \scriptsize
        (a) Content Image
    \end{minipage}
    \begin{minipage}{0.112\linewidth}
    \centering
    \scriptsize
       (b) Multimodal Input
    \end{minipage}
    \begin{minipage}{0.112\linewidth}
    \centering
    \scriptsize
        (c) MMIST \cite{Wang2024WACV}
    \end{minipage}
    \begin{minipage}{0.112\linewidth}
    \centering
    \scriptsize
       \textbf{(d) Ours}
    \end{minipage}
 \begin{minipage}{0.112\linewidth}
    \centering
    \scriptsize
        (e) Content Image
    \end{minipage}
    \begin{minipage}{0.112\linewidth}
    \centering
    \scriptsize
       (f) Multimodal Input
    \end{minipage}
    \begin{minipage}{0.112\linewidth}
    \centering
    \scriptsize
       (g) SemCS \cite{kamra2023sem}
    \end{minipage}
    \begin{minipage}{0.112\linewidth}
    \centering
    \scriptsize
        \textbf{(h) Ours} 
    \end{minipage}   
   \begin{minipage}{0.112\linewidth}
         \centering             \includegraphics[width=0.99\linewidth]{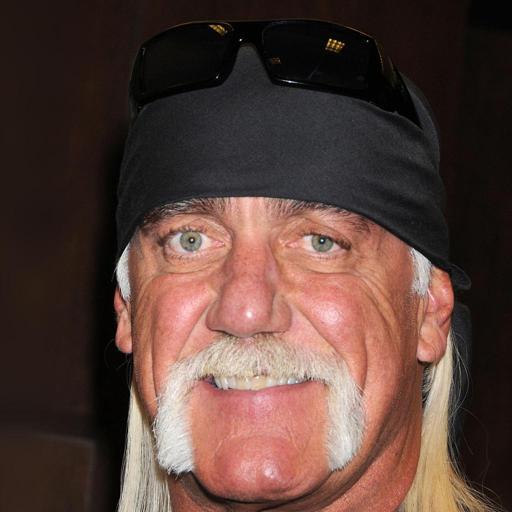}
    \end{minipage}  
    \fbox{\begin{minipage}[c][1.58cm][c]{1.58cm}
   % \begin{minipage}{0.067\linewidth}
         \centering
         \scriptsize
        Copper Plate Engraving.
    % \end{minipage}
    % \begin{minipage}{0.067\linewidth}
         \centering
             \includegraphics[height=0.60cm, width=0.60cm]{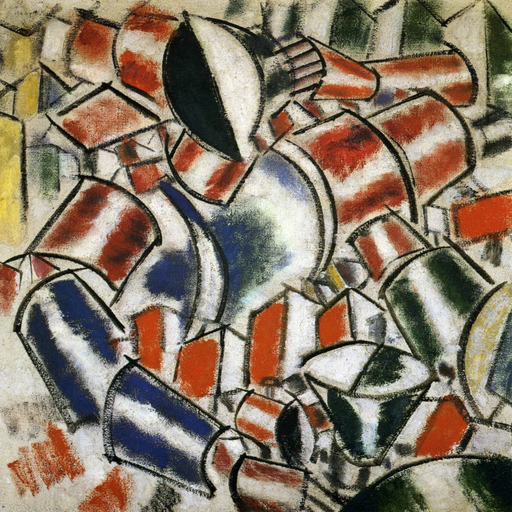}
        \end{minipage} }   
    \begin{minipage}{0.112\linewidth}
         \centering             \includegraphics[width=0.99\linewidth]{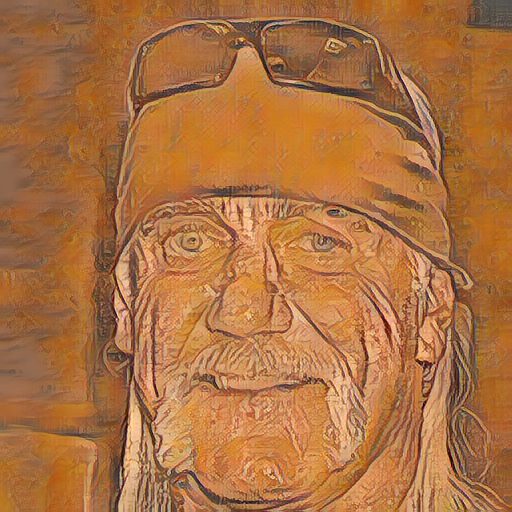}
    \end{minipage}       
    \begin{minipage}{0.112\linewidth}
         \centering             \includegraphics[width=0.99\linewidth]{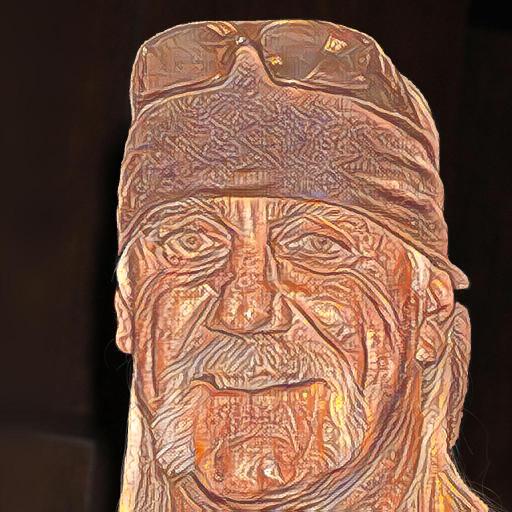}
    \end{minipage} 
      \textbf{\vline}
    \begin{minipage}{0.112\linewidth}
         \centering             \includegraphics[width=0.99\linewidth]{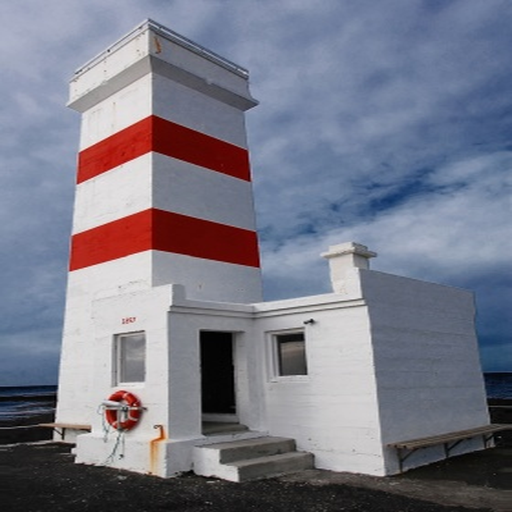}
    \end{minipage} 
   \fbox{\begin{minipage}[c][1.58cm][c]{1.58cm}
         % \centering
         \footnotesize
        \textbf{F:} Ice \\
        \textbf{B:} Starry Night by Vincent Van Gogh.
    \end{minipage}}
    \begin{minipage}{0.112\linewidth}
         \centering             \includegraphics[width=0.99\linewidth]{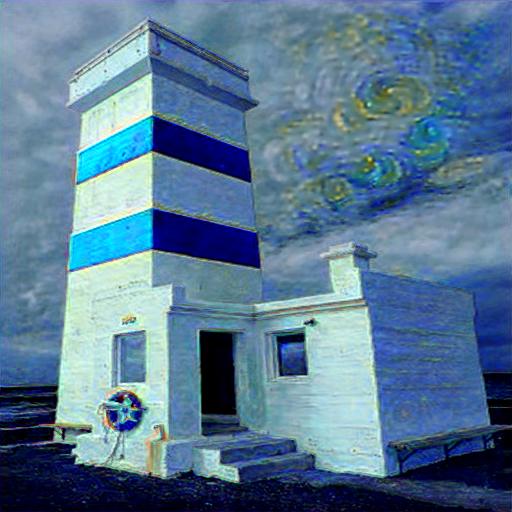}
    \end{minipage} 
    \begin{minipage}{0.112\linewidth}
         \centering             \includegraphics[width=0.99\linewidth]{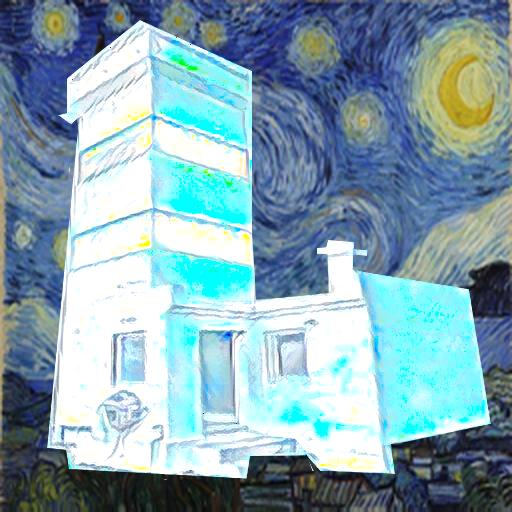}
    \end{minipage}
    % \vspace{0.2cm}
    % \vspace{0.1cm}
    \begin{minipage}{0.136\linewidth}
     \centering
     \small
        (i) Style Text
    \end{minipage}
    \begin{minipage}{0.136\linewidth}
     \centering
     \scriptsize
        (j) Content Image
    \end{minipage}
    \begin{minipage}{0.136\linewidth}
     \centering         \small (k) CS \cite{kwon2022clipstyler}
    \end{minipage}
    \begin{minipage}{0.136\linewidth}
     \centering
     \small        (l) SemCS \cite{kamra2023sem}
    \end{minipage}
    \begin{minipage}{0.136\linewidth}
     \centering
     \small        (m) CLVA \cite{fu2022language}
    \end{minipage}
    \begin{minipage}{0.136\linewidth}
     \centering
     \small       (n) MMIST \cite{Wang2024WACV}
    \end{minipage}    
    \begin{minipage}{0.136\linewidth}
     \centering
     \small
        \textbf{(o) Ours}
    \end{minipage} 
    % \vspace{-0.2cm}
    \fbox{\begin{minipage}[c][1.98cm][c]{1.98cm}
% \begin{minipage}{0.130\textwidth}
         % \centering
         \small
        A cubism style painting.
    \end{minipage}}
   \begin{minipage}{0.134\textwidth}
         \centering             \includegraphics[width=0.99\linewidth]{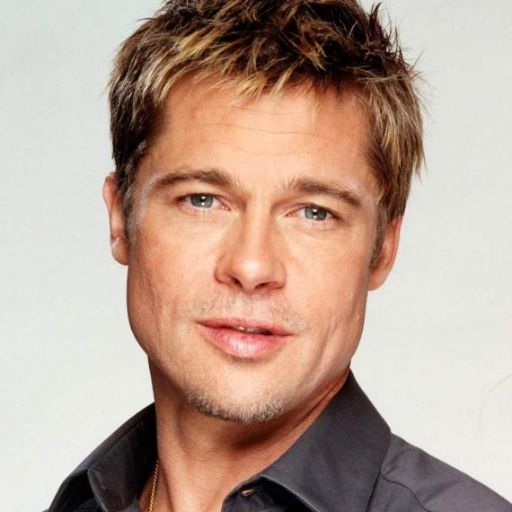}
    \end{minipage}
    \begin{minipage}{0.134\textwidth}
         \centering             \includegraphics[width=0.99\linewidth]{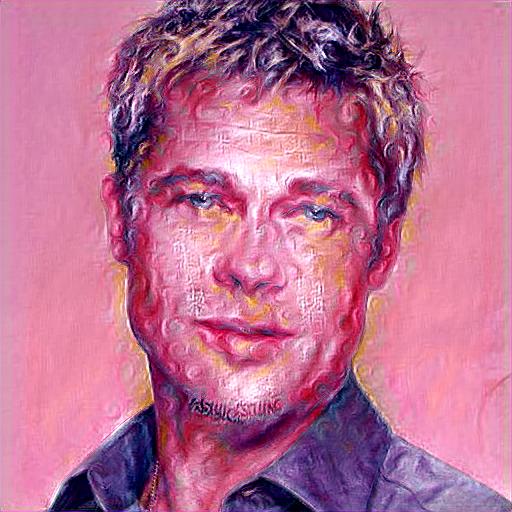}
    \end{minipage}
    \begin{minipage}{0.134\textwidth}
         \centering             \includegraphics[width=0.99\linewidth]{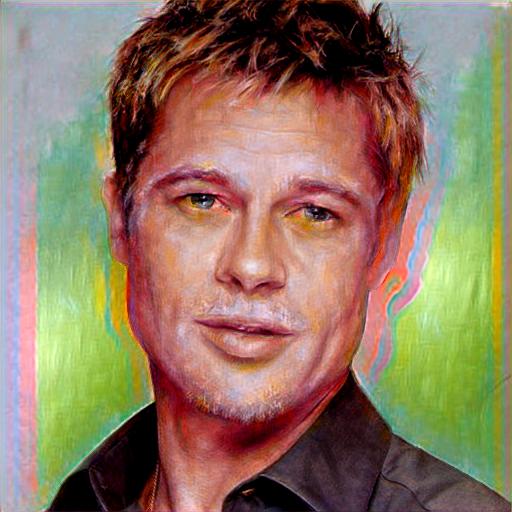}
    \end{minipage}
    \begin{minipage}{0.134\textwidth}
         \centering             \includegraphics[width=0.99\linewidth]{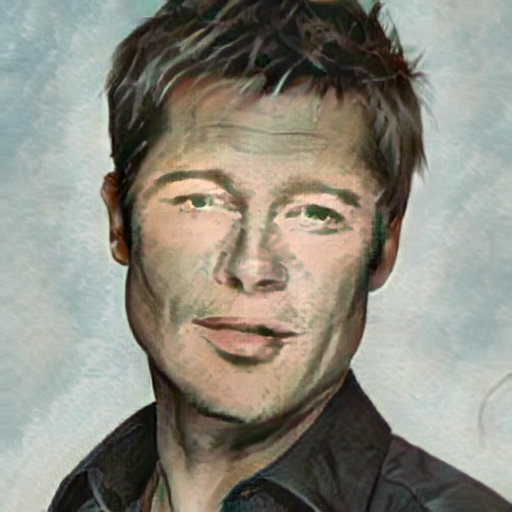}
    \end{minipage}
    \begin{minipage}{0.134\textwidth}
         \centering             \includegraphics[width=0.99\linewidth]{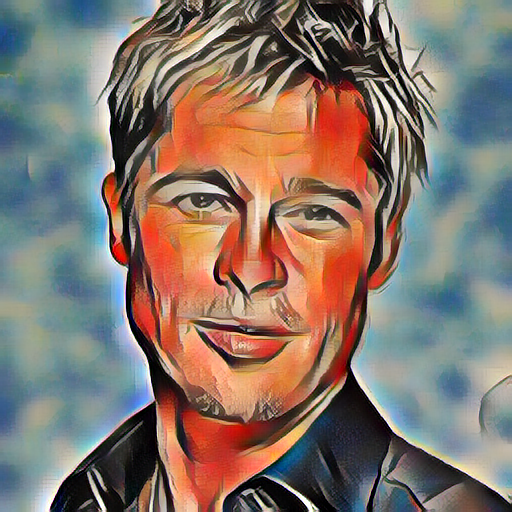}
    \end{minipage}
    \begin{minipage}{0.134\textwidth}
         \centering             \includegraphics[width=0.99\linewidth]{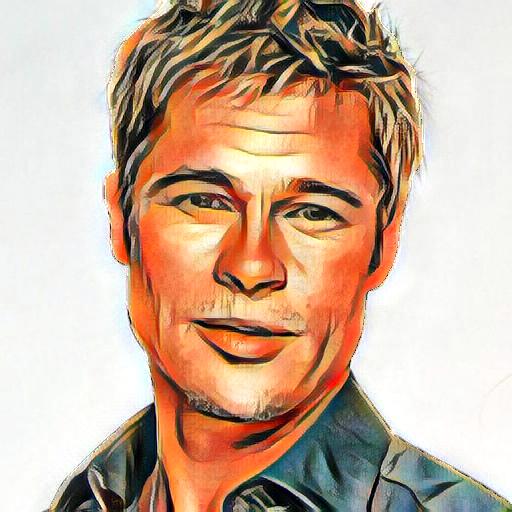}
    \end{minipage}
        % The top row presents stylized outputs (columns c-d)) obtained from multimodal style input (style text and image) applied on the salient object (car). MMIST exhibits a \textbf{spatial misalignment} problem, as copper plate features are incorrectly applied to the background. ObjMST (ours) resolves this through object-specific stylization. (f) presents foreground (F) and background (B) style input in a double text condition setting. (g) and (h) presents stylized outputs where corresponding style features are applied to the salient object (building) and surrounding elements. The second row shows the image style transfer output for single text condition
    \caption{\footnotesize The comparative stylized outputs are presented as follows: (i) Top Row, Left Side: MMIST (Single); (ii) Top Row, Right Side: TIST (Double); and (iii) Bottom Row: TIST (Single). Columns (a, e, j) represent the content images, while columns (b, f, i) show the multimodal style inputs. In MMIST~\cite{Wang2024WACV} (column c), misalignment is evident as the texture and color of the copper plate features are inconsistent compared to Ours-ObjMST (column d). In TIST (Double), SemCS~\cite{kamra2023sem} (column g) introduces undesired distortions, whereas ObjMST (column h) correctly applies the "Starry Night" style features to the background (sky) and ice features to the foreground. In TIST (Single), ObjMST (column o) effectively preserves the content features while accurately applying the desired style.}
     \label{fig:intro}
\end{figure*}

 \begin{table}[]
\centering
\scriptsize
\caption{\footnotesize The table compares Image Style Transfer (IST) methods. 'Single' in columns one and three indicates a single text condition, while 'double' in column two refers to double text conditions for foreground \textbf{($fg$)} and background \textbf{($bg$)} objects.}
\begin{tabular}{|l|c|c|c|}
\hline
                                                                   & \textbf{\begin{tabular}[c]{@{}c@{}}Text-Based IST\\ (Single)\end{tabular}} & \textbf{\begin{tabular}[c]{@{}c@{}}Text-Based IST\\ (Double)\end{tabular}} & \textbf{\begin{tabular}[c]{@{}c@{}}Multimodal IST\\ (Single)\end{tabular}} \\ \hline
\textbf{CS} \cite{kwon2022clipstyler}                                                      & $\surd$                                                                    & X                                                                          & X                                                                          \\ \hline
\textbf{SemCS} \cite{kamra2023sem}                                                     & $\surd$                                                                    & $\surd$                                                                    & X                                                                          \\ \hline
\textbf{LDAST} \cite{fu2022language}                                                    & $\surd$                                                                    & X                                                                          & X                                                                          \\ \hline
\textbf{MMIST} \cite{Wang2024WACV}                                                    & $\surd$                                                                    & X                                                                          & $\surd$                                                                    \\ \hline
\textbf{\begin{tabular}[c]{@{}l@{}}ObjMST \\  (Ours)\end{tabular}} & $\surd$                                                                    & $\surd$                                                                    & $\surd$                                                                    \\ \hline
\end{tabular}
\label{tab:style_transfer}
\end{table}
We observed that MMIST~\cite{Wang2024WACV} encounters difficulties in preserving semantics in the style transfer. For instance, Fig.~\ref{fig:intro}, top row (a-d) shows that MMIST~\cite{Wang2024WACV} fails to capture color and texture tones corresponding to the style-text description "Copper plate engraving". This limitation is primarily due to the difficulty in aligning text and image~\cite{neucomspatial22yu} features within the shared CLIP (Contrastive Language-Image Pre-training) \cite{radford2021learning} embedding space.
% We, using ObjMST, are able to generate style features consistent with the input style text-image pair.

Another issue in MMIST~\cite{Wang2024WACV} is content mismatch, where a similar style is applied to semantically different objects, resulting in unintended spill-over. For example,  Fig.~\ref{fig:intro} (e-f) shows that the texture or patterns from the style features spill over into regions such as the background or other objects, producing a visually incoherent result in MMIST~\cite{Wang2024WACV}. SemCS~\cite{kamra2023sem} and  CLVA \cite{fu2022language} were also observed to be suffering from content mismatch (Fig.~\ref{fig:intro}). 

We address the issue of misalignment in style transfer by introducing ObjMST, the first object-focused multimodal style transfer framework. The main challenge lies in performing StyleGAN inversion, guided by directional CLIP loss, to generate intermediate style representations. These representations are then used in image-based style transfer to produce the final output. Our analysis of the intermediate style representations (Table~\ref{tab:sty_reps} and Fig.~\ref{fig:ablation1}) demonstrates improved visual quality when both salient objects and their surrounding elements are considered. To further enhance the representations of salient objects, we mask out arbitrary content features from the multimodal style input during the StyleGAN inversion process. This ensures the preservation of essential style elements in the intermediate stages. This entire process leads to the development of the "masked" directional CLIP loss, which ensures consistent and aligned generation of intermediate style representations.

ObjMST resolves the content mismatch between salient objects and surrounding elements by employing salient-to-key feature matching \cite{zhu2023all}. The alignment of each salient object's content features with stable key positions of style features (S2K) facilitates semantically preserved stylization, focusing exclusively on the salient object within the image.  For example, in Fig.~\ref{fig:intro}-(h), ice features are applied solely to the building.

ObjMST generates distinct style representations for salient objects and surrounding elements using cross-model GAN inversion, assisted by masked directional CLIP loss for the foreground ($fg$) and background ($bg$) objects. Separating the style representation generation for $fg$ and $bg$ is essential to minimizing content mismatches. To ensure consistency in the final output, self-supervised image harmonization is employed~\cite{jiang2021ssh}. Our major contributions are as follows:

\begin{itemize}[noitemsep, left=0pt]
    \item We propose the ObjMST framework, which provides separate style supervision for salient objects and surrounding elements, while addressing alignment issues in multimodal representation learning (Fig.~\ref{fig:multimodal} and Fig.~\ref{fig:double_style_res}).
    \item The ObjMST framework maps salient content features to stable style key features \textit{(S2K)} to preserve the semantic structure of content features (Fig.~\ref{fig:outputs1} and Table~\ref{table:scores}). 
      \item We validate experimental results rigorously using Contrique~\cite{madhusudana2022image}, NIMA~\cite{talebi2018nima} scores, and user studies (Table~\ref{table:scores}).
\end{itemize}
\begin{figure*}[!htb]
    \centering
    \includegraphics[width=0.95\textwidth, height=8.5cm]{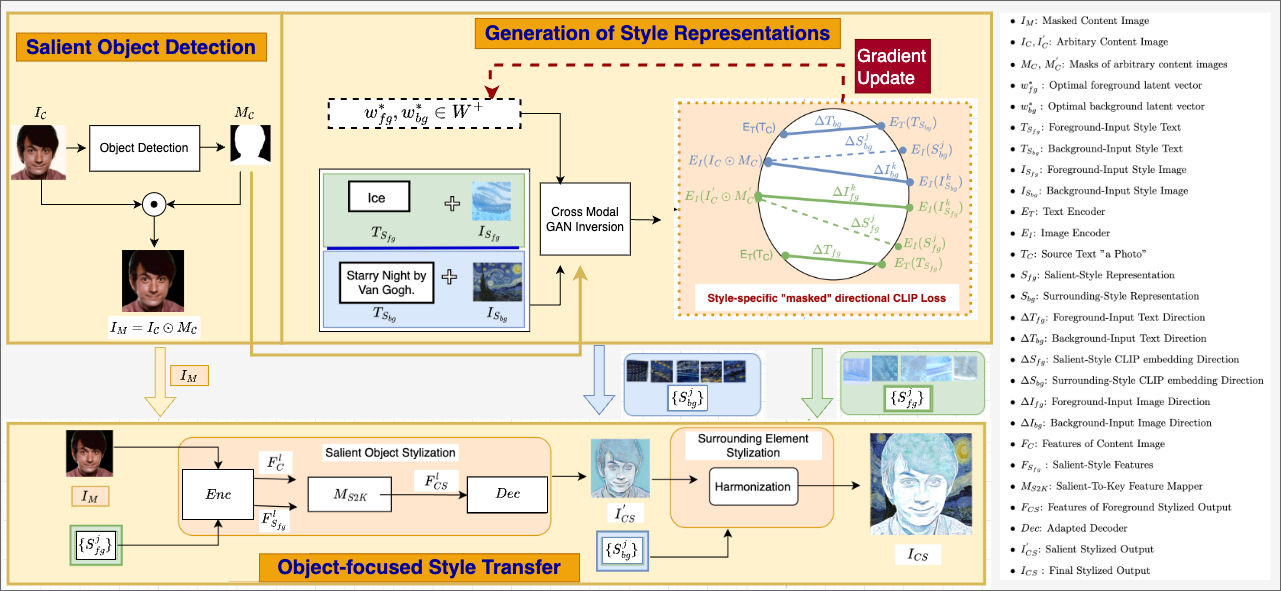}
    % \scriptsize  
    % multimodal input of foreground style text-image pair ($T_{S_{fg}} \text{,} I_{S_{fg}}$) and background style text-image pair ($T_{S_{bg}} \text{,} I_{S_{bg}}$) is passed to cross-modal GAN Inversion.  
    
    \caption{\footnotesize The figure illustrates the proposed ObjMST framework. Given the segmentation mask ($M_{\mathcal{C}}$) of the content image ($I_{\mathcal{C}}$), we compute the masked content image ($I_M$). In Step~1, we compute the optimal foreground and background latent vector $w_{fg}^*$ and $w_{bg}^*$ to obtain the salient and surrounding style representations $S_{fg}$ and $S_{bg}$. This is achieved by passing multimodal input of foreground-input style text-image pair ($T_{S_{fg}} \text{,} I_{S_{fg}}$) and background-input style text-image pair ($T_{S_{bg}} \text{,} I_{S_{bg}}$) to cross-modal GAN Inversion, which is trained using the proposed masked directional Style CLIP Loss ($L_{fg}$). In Step~2, the foreground stylized output ($I_{CS}^{'}$) is generated by mapping salient content features ($F_C^l$) to stable style key features ($F_{S_{fg}}^l$) through ($M_{S2K}$) mapper. Finally,  surrounding-style ($S_{bg}$) representation is applied to the background through image harmonization to generate stylized output $I_{CS}$.}
    \label{fig:block_diagram}
\end{figure*}
\vspace{-0.6cm}
\section{Related Work}
\label{sec:relatedWork}
\noindent \textbf{Arbitrary Style Transfer.} Arbitrary style transfer (AST) methods \cite{DBLPBatziouIPVK23, yao2019attention, liu2021adaattn} have gained attention for their effectiveness in transferring arbitrary styles. Local feature distribution methods often employ attention-based feature matching \cite{yao2019attention, liu2021adaattn}. Adaattn \cite{liu2021adaattn} introduced dense correspondence via all-to-all attention \cite{liu2021adaattn} but it faced challenges with distorted patterns and unstable matching. The All-to-Key attention mechanism \cite{zhu2023all} addresses these issues by identifying stable style keys through distribution and progressive attention.\\~
% \vspace{-0.0cm}
\noindent \textbf{Multimodality-Guided Image Style Transfer.} Text-guided style transfer \cite{kwon2022clipstyler, kamra2023sem, fu2022language, Wang2024WACV} has gained attraction by eliminating the need for a style image. However, methods such as CLIPStyler (CS) \cite{kwon2022clipstyler} and CLVA \cite{fu2022language} encounter issues with spatial misalignment and overstylization \cite{kamra2023sem}. SemCS \cite{kamra2023sem} addresses these problems through controllable transfer using global foreground and background loss, while MMIST \cite{Wang2024WACV}, the first multimodal style transfer method, also suffers from misalignment. \\~
\noindent \textbf{CLIP and StyleGAN.} CLIP plays a significant role in multimodal contexts, bridging the gap between visual and textual information. StyleCLIP \cite{patashnik2021styleclip} introduced global CLIP loss to minimize cosine distance between the generated image and target text, while StyleGAN-Nada \cite{gal2022stylegan} employed local directional loss for better diversity. Images are either encoded or inverted into latent space for editing via latent vector manipulation \cite{DBLP:journals/tip/MaoWWWCJM24}. Recent models use CLIP \cite{radford2021learning} embeddings to guide text-based editing \cite{gal2022stylegan, kwon2022clipstyler, patashnik2021styleclip}.

\vspace{-0.3cm}
\section{Methodology} \label{sec:approach}
We illustrate Object-focused Multimodal Style-transfer framework (ObjMST) in Fig.~\ref{fig:block_diagram}. It comprises of two steps: generating style representations and performing object-focused style transfer. Both steps require the segmentation mask of the content image as described in the below subsections. We consider the Segment Anything method \cite{Kirillov2023ICCV} to obtain the segmentation mask of the content image.
\subsection{Generation of Style Representations}
\label{sec:gen_styreps}
We generate two distinct style representations for the foreground ($fg$) and background ($bg$), referred to as the salient and surrounding style representations. \\~
% Both representations require distinct arbitrary content image $I_{\mathcal{C}}$ and $I_{\mathcal{C}}^{'}$ and their corresponding masks $M_{\mathcal{C}}$ and $M_{\mathcal{C}}^{'}$ to ensure that these representations are not biased by the features of the specific arbitrary content image.\\~
\noindent \textbf{Salient-Style Representations.}
We propose style-specific "\textit{masked}" directional CLIP loss to synthesize consistent salient-style representations $S_{fg}$ as shown in Fig.~2. First, we pass foreground-input style text-image pairs ($T_{S_{fg}} \text{, } I_{S_{fg}}$) to cross-modal GAN inversion. Next, the latent vector $w_{fg}$ is passed to StyleGAN3~\cite{karras2021alias} generator $G$ with random initialization to obtain $S_{fg}=G(w_{fg})$. Subsequently, the foreground-input style image ($I_{S_{fg}}$) and salient-style representations ($S_{fg}$) undergo cropping and augmentation to produce patch-level representations $I_{S_{fg}}^k$ and $S_{fg}^j$, respectively, where j and k denote the indices of the number of patches ($N_{crop}$). \\~
% the cosine similarity on two pairs of CLIP embedding directions: The foreground-input style image embedding direction ($\Delta I_{fg}^k$) and salient style representation embedding direction ($\Delta S_{fg}^j$).
\noindent The directional CLIP loss used in MMIST~\cite{Wang2024WACV} generates imprecise style representations due to the misalignment between input image and text style features in the shared CLIP embedding space. To address this issue, we introduce a masked directional CLIP loss, which excludes irrelevant content features from CLIP space. \\~
The first part of the proposed loss function involves computing the
average cosine similarity between CLIP embedding directions of the patched image-image pair, specifically the foreground-input style image, ($I^k_{S_{fg}}$) and salient-style representations, ($S_{fg}^j$). 
% Their corresponding direction pairs are foreground-input and salient style embedding directions $\Delta I_{fg}^k$ and $\Delta S_{fg}^j$, respectively.
The computation of foreground-input CLIP embedding direction ($\Delta I_{fg}^k$) explicitly masks out the arbitrary content features ($I_{C} \odot M_{C}$) from $I_{fg}^k$ as described in the equation below: 
\vspace{-0.2cm}
\begin{equation}
    \Delta I^k_{fg} = E_I(I^k_{S_{fg}}) - E_I (I_{C} \odot M_{C}) 
    \label{eq:eq1}
\end{equation}
\vspace{-0.1cm}
Here, $E_I$ denotes the image encoder. Similarly, salient-style CLIP embedding direction ($\Delta S_{fg}^j$) removes arbitrary content features from the salient-style representations as shown below:
% through hadamard product. 
\vspace{-0.3cm}
\begin{equation}
    \Delta S^j_{fg} = E_I(S^j_{fg}) - E_I (I_{C} \odot M_{C})    
    \label{eq:eq2}
\end{equation}
\noindent The second part of the proposed loss function involves computing average cosine similarity between CLIP embedding directions of image-text pair, specifically foreground-input style text, ($T_{S_{fg}}$) and the patched salient-style representation ($S_{fg}^j$). The CLIP embedding direction of foreground-input style text $\Delta T_{fg}$ is defined as the difference between foreground-input style text embeddings $T_{S_{fg}}$ and source text embeddings ($T_{C}$) i.e $ \Delta T_{fg} = E_T(T_{S_{fg}}) - E_T(T_C)$. Here, $T_C$ represents the style text for any natural image and is set to "a photo" ~\cite{kwon2022clipstyler}, while $E_T$ denotes text encoder. The masked directional CLIP loss is defined as the averaged cosine similarity for each pair of CLIP embedding directions (image-text and image-image) as follows: 
% . follows:
% which is the averaged cosine similarity for each pair of CLIP embedding directions (image-text and image-image). 
\begin{equation} 
\begin{split}
L_{fg} = \frac{1}{N_{\text{crop}}} \sum_{j=1}^{N_{\text{crop}}}  \left(1 - \frac{\Delta S^j_{fg} \cdot \Delta T_{fg}}{\|\Delta S^j_{fg}\| \|\Delta T_{fg}\|}\right)  +  \\ 
\lambda * \frac{1}{N^2_{\text{crop}}}\sum_{j=1}^{N_{\text{crop}}}\sum_{k=1}^{N_{\text{crop}}} \left(1 - \frac{\Delta S^j_{fg} \cdot \Delta I^{k}_{fg}}{\|\Delta S^j_{fg}\| \|\Delta I^{k}_{fg}\|}\right)    
\end{split}
\label{eq:Loss_fg}
\end{equation}
Here, $\lambda$ is the tunable parameter. We minimize the $L_{fg}$ to find the optimal foreground latent vector ($w_{fg}^*$), where $w_{fg}^* = argmin_w L_{fg}$. Finally, we pass the foreground optimal vector ($w_{fg}^*$) to the generator $G$ to obtain the salient-style representations, i.e., $S_{fg} = G(w_{fg}^{*})$. \\~
\noindent \textbf{Surrounding-Style Representations.}
 Similarly to salient-style representations, we determine the optimal background latent vector ($w_{bg}^*$) to generate surrounding-style representations $S_{bg}$.  To achieve this, the background-input style text-image pair ($T_{S_{bg}} \text{, } I_{S_{bg}}$) is used as input. As illustrated in Fig.~\ref{fig:block_diagram}, we subtract the Hadamard product of another randomly chosen content image $I_{C}^{'}$ and its segmentation mask $M_{C}^{'}$ from the patched background-input and surrounding-style representations ($I_{S_{bg}}^k$ and $S_{bg}^j$) to compute their corresponding CLIP embedding directions ($\Delta I_{bg}^k$ and $\Delta S_{bg}^j$). To ensure that these representations are not biased by features of a specific arbitrary content image, another arbitrary content image $I_{C}^{'}$ and its segmentation mask $M_{C}^{'}$ are chosen. Thus, the background-input CLIP embedding direction ($\Delta I_{bg}^k$) is formulated as follows:
\begin{equation}
\Delta I^k_{bg} = E_I(I^k_{S_{bg}}) - E_I (I_{C}^{'} \odot M_{C}^{'}) 
 \label{eq:eq4}
\end{equation}
Similarly, surrounding-style CLIP embedding direction ($\Delta S_{bg}^j$) is defined below:  
\begin{equation}
\Delta S^j_{bg} = E_I(S^j_{bg}) - E_I (I_{C}^{'} \odot M_{C}^{'}) 
 \label{eq:eq5}
\end{equation}
The background ($w_{bg}^{*}$) latent vector for surrounding-style representations is optimized by minimizing $L_{bg}$, i.e. $w_{bg}^* = argmin_w L_{bg}$, where $L_{bg}$ is defined as follows:
\begin{equation} 
\begin{split}
  L_{bg} = \frac{1}{N_{\text{crop}}} \sum_{j=1}^{N_{\text{crop}}}  \left(1 - \frac{\Delta S^j_{bg} \cdot \Delta T_{bg}}{\|\Delta S^j_{bg}\| \|\Delta T_{bg}\|}\right)  +  \\
\lambda * \frac{1}{N^2_{\text{crop}}}\sum_{j=1}^{N_{\text{crop}}}\sum_{k=1}^{N_{\text{crop}}} \left(1 - \frac{\Delta S^j_{bg} \cdot \Delta I^{k}_{bg}}{\|\Delta S^j_{bg}\| \|\Delta I^{k}_{bg}\|}\right)  
\end{split}
\label{eq:Loss_bg}
\end{equation}
% \textcolor{red}{Next Check...}
 Finally, the surrounding-style representation $S_{bg}$ is obtained by passing the optimal background vector to Stylegan3~\cite{karras2021alias} generator $G$, i.e., $S_{bg} = G(w_{bg}^{*})$. Incorporating the masked component in the computation of CLIP embedding direction ensures meaningful alignment between the generated style representations and multimodal inputs. \\~
\vspace{-0.4cm}
\subsection{Object Stylization}
\label{subsec:ObjST}
The bottom part of Fig.~\ref{fig:block_diagram} illustrates the object-focused style transfer step. The goal is to use salient-style representations $S_{fg}$ to stylize the salient object of the content image $I_{\mathcal{C}}$ and use surrounding-style representations $S_{bg}$ to stylize the background part of the content image $I_{\mathcal{C}}$ .\\~
\noindent \textbf{Salient Object Stylization.} 
We extract content features ($F_C$) from the masked content image ($I_M=I_C \odot M_C$) and style features ($F_{S_{fg}}$) from salient-style representations $S_{fg}$, (obtained in Sec.~\ref{sec:gen_styreps}). To obtain the multi-scale content $F_C^l$ and style features $F_{S_{fg}}^l$ at different layers $l$  such as $Relu\{3\_1, 4\_1, 5\_1\}$, we pass $I_M$ and patched salient-style representations $S_{fg}^j$ through a pretrained VGG19~\cite{simonyan2015a} encoder ($Enc$) with fixed parameters. Here, $F \in R^{C_l*H_l*W_l}$, where $C_l$, $H_l$ and $W_l$ are the number of channels, height and width of image at different layers $l$. The feature extraction of $F_C^l$ and $F_{S_{fg}}^l$ is described below:
\begin{equation}
F^l_C = Enc(I_C \odot M_C), \quad F^l_{S_{fg}} = Enc({S_{fg}^j})
\label{eq:eq7}
\end{equation}
% The segmentation mask $M_{\mathcal{C}}$ of content image $I_{\mathcal{C}}$ is obtained by SAM~\cite{Kirillov2023ICCV}.
% , $I_{\mathcal{C}} \odot M_{\mathcal{C}}$
% (Sec.~\ref{sec:gen_styreps}). This involves obtaining multi-scale features of the masked content image, $I_{\mathcal{C}} \odot M_{\mathcal{C}}$ ($M_{\mathcal{C}}$ is obtained through SAM~\cite{Kirillov2023ICCV}) and intermediate style representations $S_{fg}$ using pre-trained VGG-19~\cite{simonyan2015a} ($Enc$) as shown in Fig. \ref{fig:block_diagram} with their fixed parameters.
% Here, $l$ denotes extracted feature at $Relu\{3\_1, 4\_1, 5\_1\}$ 
% layers and $F \in R^{C_l*H_l*W_l}$. 
% Content features are excluded to ensure that style features are applied only to prominent objects.
 Motivated by attention-based arbitrary style transfer methods \cite{liu2021adaattn, zhu2023all}, we employ all-to-key \cite{zhu2023all} mapping on salient objects features to perform Salient-To-Key (S2K) feature mapping. This introduces Salient-To-Key feature mapper $M_{S2K}$ to generate transformed multi-scale features of foreground stylized object $F^{l}_{CS}$ as:  
 $F^{l}_{CS} = M_{S2K}(F^l_C, F^l_{S_{fg}})$.
 $M_{S2K}$ module maps salient-style features ($F_{S_{fg}}^l$) with features of content image ($F_C^l$) while preserving content integrity through distributive and progressive attention \cite{zhu2023all}. 
 % This is achieved by mapping salient object features from the content image to stable key features of the salient-style representations. 
 Finally, the salient stylized output \( I_{CS}^{'} \) is reconstructed from multi-scale transferred feature of stylized output\( \{F^{l}_{CS}\}\) by passing them through a decoder ($Dec$). We adopted the decoder ($Dec$) from AdaAttn \cite{liu2021adaattn}, and is described as $I_{CS}^{'} = Dec(\{F^l_{CS}\})$. \\~
 % instead of all-to-all ~\cite{liu2021adaattn} key mapping
 % The extracted features are passed to the Salient-to-Key feature mapper ($M_{S2K}$), generating feature maps for foreground stylized output ($F_{CS}^{'}$).
% In this equation, \( F^l_{CS} \) represents the transformed features at layer l, obtained by applying the Saliency-to-Key attention module 
% \( M^l_{S2K} \) to content features \( F^l_C \) and salient style features \( F^l_{S_{fg}} \). 

% \[ I_{CS}^{'} = Dec(\{F^l_{CS}\}) \] 
% \vspace{-0.4cm}
\noindent \textbf{Surrounding Element Stylization.}
To stylize the background of content image, we realistically blend the surrounding-style representations $\{S_{bg}^j\}$ with the foreground stylized output ($I_{CS}^{'}$). We utilized the self-supervised Harmonization framework (SSH \cite{jiang2021ssh}), which is trained on arbitrary unedited content images. In this way, we obtain the final stylized output $I_{CS}$, where distinct style features applied separately to foreground and background.
\begin{figure*}[!htb]
    \centering
    \begin{minipage}{0.130\linewidth}
     \centering
     \small
        a) Style Text
    \end{minipage}
    \begin{minipage}{0.130\linewidth}
     \centering
     \small
        b) Content Image
    \end{minipage}
    \begin{minipage}{0.130\linewidth}
     \centering         \small c) CS\cite{kwon2022clipstyler}
    \end{minipage}
    \begin{minipage}{0.130\linewidth}
     \centering
     \small        d) SemCS\cite{kamra2023sem}
    \end{minipage}
    \begin{minipage}{0.130\linewidth}
     \centering
     \small        e) CLVA\cite{fu2022language}
    \end{minipage}
    \begin{minipage}{0.130\linewidth}
     \centering
     \small       f) MMIST\cite{Wang2024WACV}
    \end{minipage}    
    \begin{minipage}{0.130\linewidth}
     \centering
     \small
        \textbf{g) Ours}
    \end{minipage} 
        \begin{minipage}{0.130\textwidth}
         % \centering
         \small
        Desert Sand
    \end{minipage}
     \begin{minipage}{0.130\textwidth}
         \centering             \includegraphics[width=0.99\linewidth]{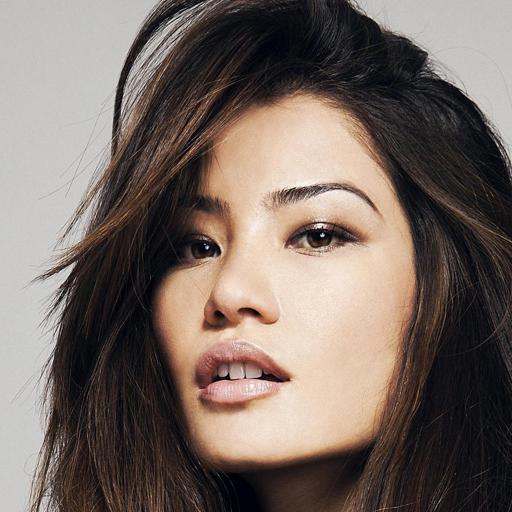}
    \end{minipage}
    \begin{minipage}{0.130\textwidth}
         \centering             \includegraphics[width=0.99\linewidth]{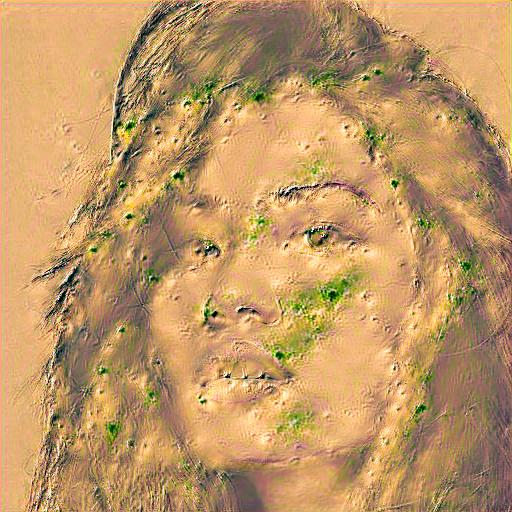}
    \end{minipage}
    \begin{minipage}{0.130\textwidth}
         \centering             \includegraphics[width=0.99\linewidth]{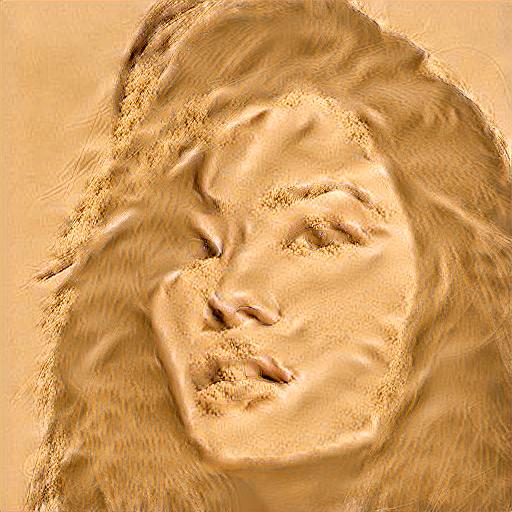}
    \end{minipage}
    \begin{minipage}{0.130\textwidth}
         \centering             \includegraphics[width=0.99\linewidth]{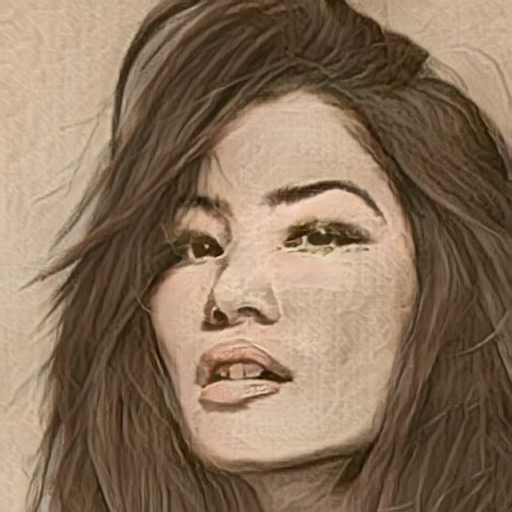}
    \end{minipage}
    \begin{minipage}{0.130\textwidth}
         \centering             \includegraphics[width=0.99\linewidth]{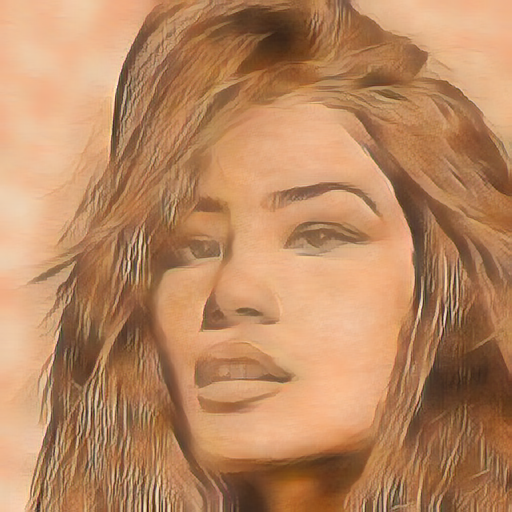}
    \end{minipage}
    \begin{minipage}{0.130\textwidth}
         \centering             \includegraphics[width=0.99\linewidth]{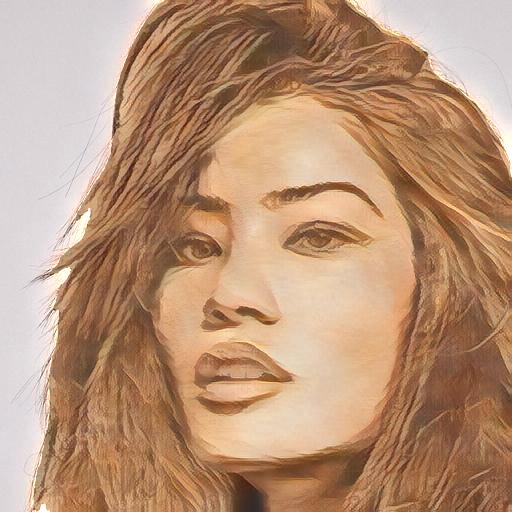}
    \end{minipage}
      \begin{minipage}{0.130\textwidth}
         % \centering
         \small
        Green Crystal
    \end{minipage}
     \begin{minipage}{0.130\textwidth}
         \centering             \includegraphics[width=0.99\linewidth]{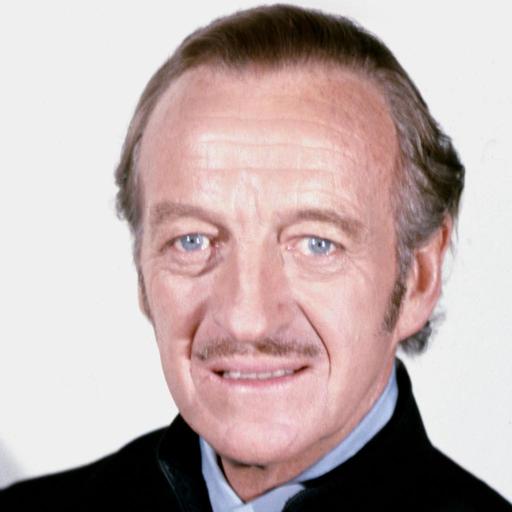}
    \end{minipage}
    \begin{minipage}{0.130\textwidth}
         \centering             \includegraphics[width=0.99\linewidth]{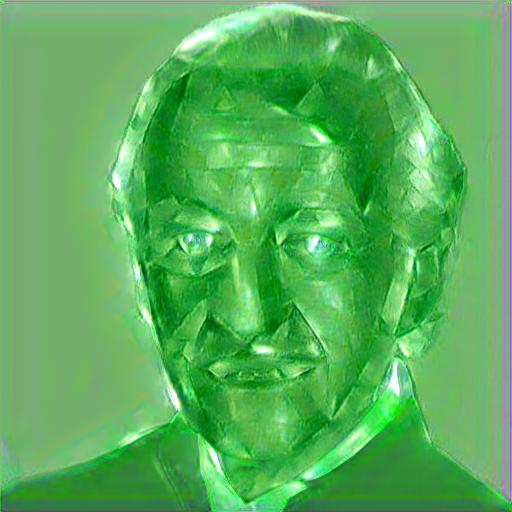}
    \end{minipage}
    \begin{minipage}{0.130\textwidth}
         \centering             \includegraphics[width=0.99\linewidth]{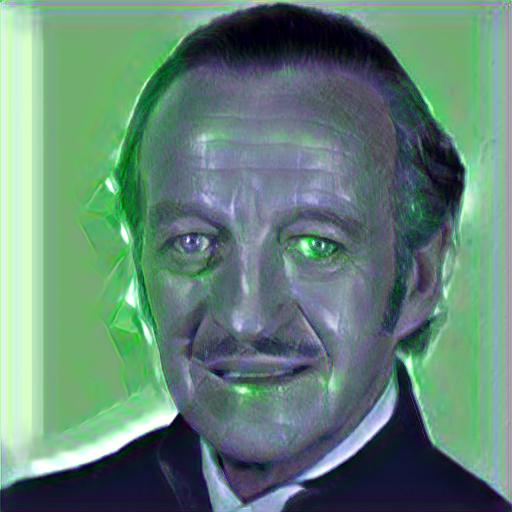}
    \end{minipage}
    \begin{minipage}{0.130\textwidth}
         \centering             \includegraphics[width=0.99\linewidth]{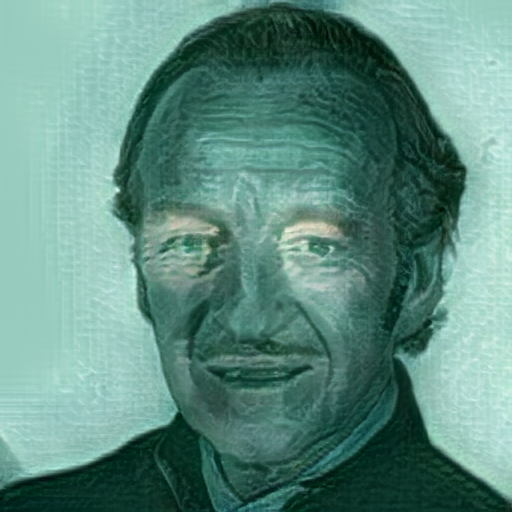}
    \end{minipage}
    \begin{minipage}{0.130\textwidth}
         \centering             \includegraphics[width=0.99\linewidth]{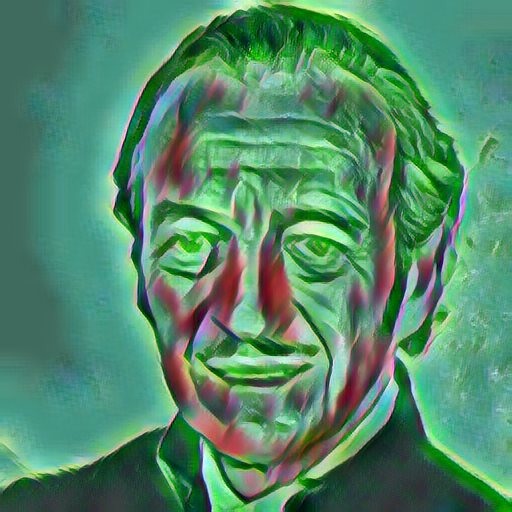}
    \end{minipage}
    \begin{minipage}{0.130\textwidth}
         \centering             \includegraphics[width=0.99\linewidth]{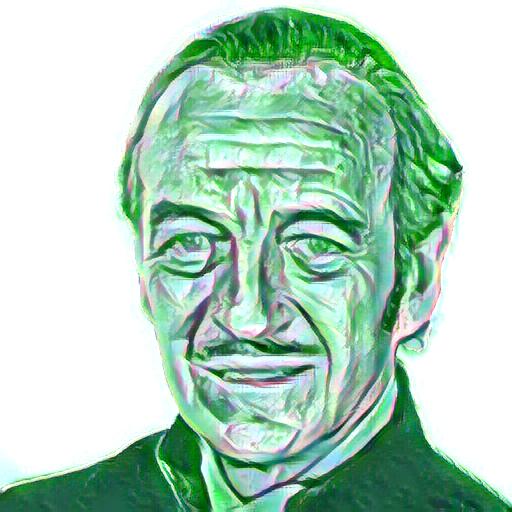}
    \end{minipage}  
    % The figure shows text-based image style transfer with single text condition.
 \caption{\footnotesize \textbf{Text-based IST (Single).} ObjMST (g) better preserves the facial structure and harmoniously integrates text and visual style cues, particularly in terms of texture and color consistency, compared to the baseline methods (c-f).} 
\label{fig:outputs1}
\end{figure*}
\begin{figure*}
   \begin{minipage}{0.128\linewidth}
     \centering
     \small
        a) Style Text
    \end{minipage}
    \begin{minipage}{0.128\linewidth}
     \centering
     \small
        b) Content Image
    \end{minipage}
    \begin{minipage}{0.128\linewidth}
     \centering         \small c) CS \cite{kwon2022clipstyler}
    \end{minipage}
    \begin{minipage}{0.128\linewidth}
     \centering
     \small        d) SemCS \cite{kamra2023sem}
    \end{minipage}
    \begin{minipage}{0.128\linewidth}
     \centering
     \small        e) CLVA \cite{fu2022language}
    \end{minipage}
    \begin{minipage}{0.128\linewidth}
     \centering
     \small       f) MMIST \cite{Wang2024WACV}
    \end{minipage}    
    \begin{minipage}{0.128\linewidth}
     \centering
     \small
        \textbf{g) Ours}
    \end{minipage}     
     \begin{minipage}{0.131\linewidth}
         \centering
         \small
         Fire
    \end{minipage}
     \begin{minipage}{0.131\linewidth}
         \centering             \includegraphics[width=0.84\linewidth]{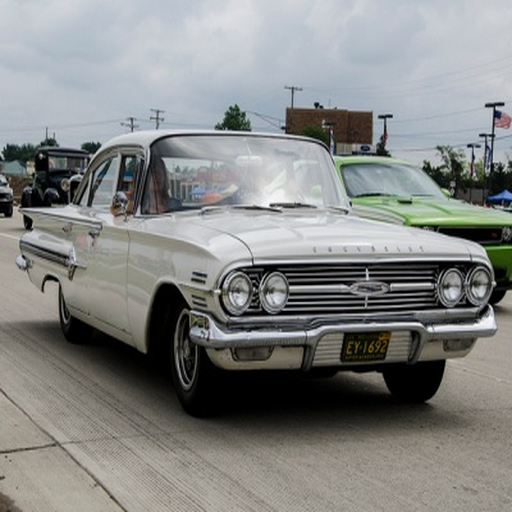}
    \end{minipage}
    \begin{minipage}{0.131\linewidth}
         \centering             \includegraphics[width=0.84\linewidth]{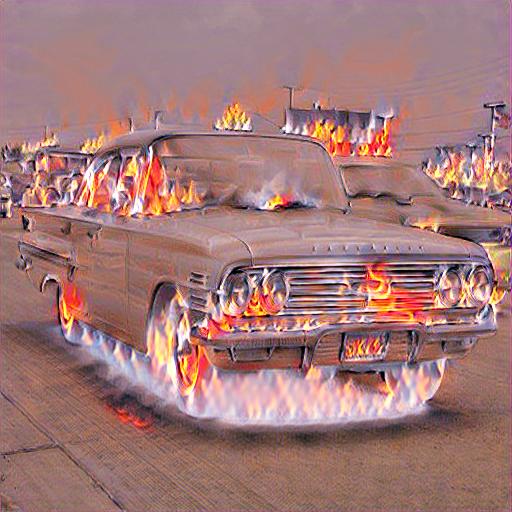}
    \end{minipage}
    \begin{minipage}{0.131\linewidth}
         \centering             \includegraphics[width=0.84\linewidth]{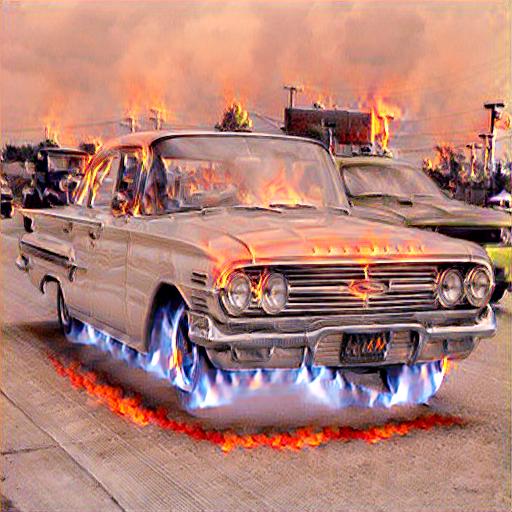}
    \end{minipage}
    \begin{minipage}{0.131\linewidth}
         \centering             \includegraphics[width=0.84\linewidth]{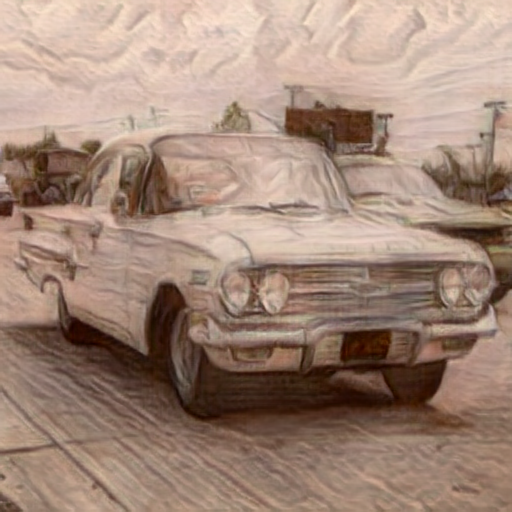}
    \end{minipage}
    \begin{minipage}{0.131\linewidth}
         \centering             \includegraphics[width=0.84\linewidth]{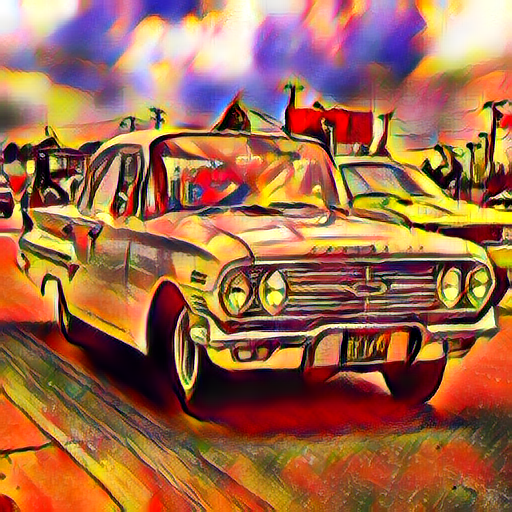}
    \end{minipage}
    \begin{minipage}{0.131\linewidth}
         \centering             \includegraphics[width=0.84\linewidth]{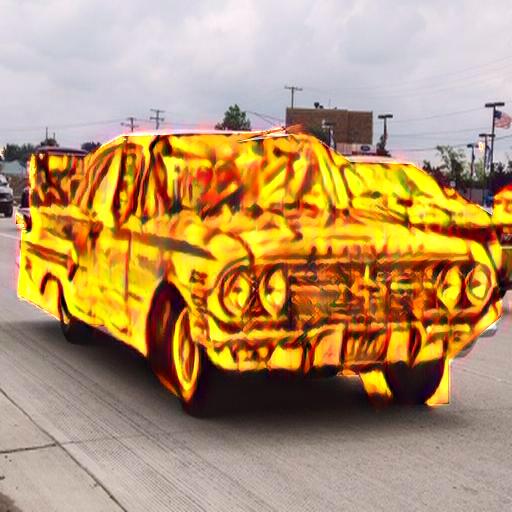}
    \end{minipage}
     \begin{minipage}{0.131\linewidth}
         \centering
         \small
        A graffiti style painting
    \end{minipage}
     \begin{minipage}{0.131\linewidth}
         \centering             \includegraphics[width=0.84\linewidth]{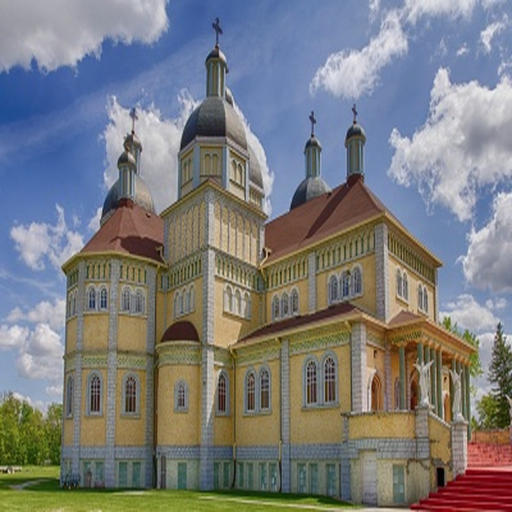}
    \end{minipage}
    \begin{minipage}{0.131\linewidth}
         \centering             \includegraphics[width=0.84\linewidth]{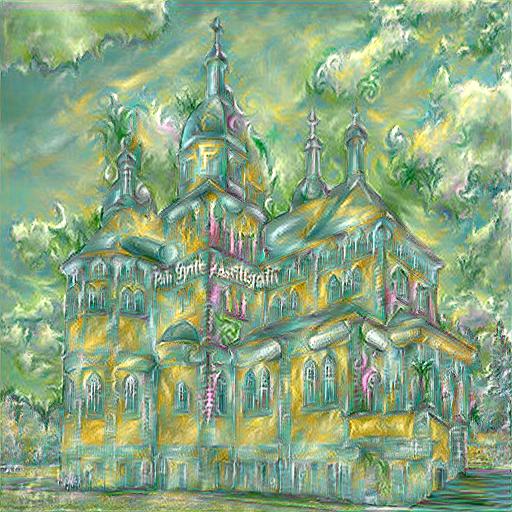}
    \end{minipage}
    \begin{minipage}{0.131\linewidth}
         \centering             \includegraphics[width=0.84\linewidth]{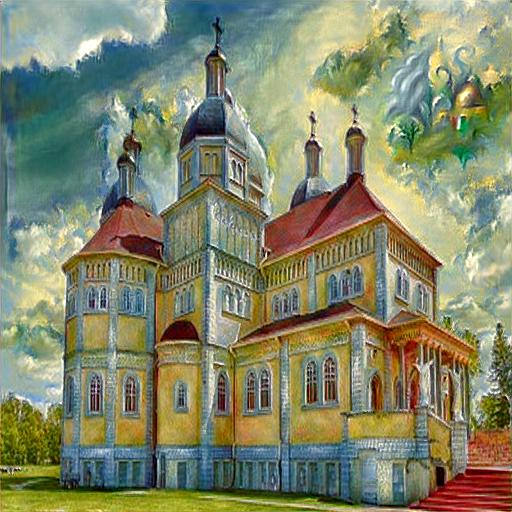}
    \end{minipage}
    \begin{minipage}{0.131\linewidth}
         \centering             \includegraphics[width=0.84\linewidth]{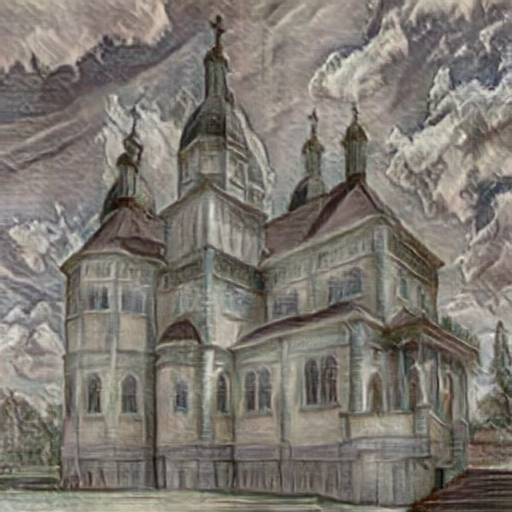}
    \end{minipage}
    \begin{minipage}{0.131\linewidth}
         \centering             \includegraphics[width=0.84\linewidth]{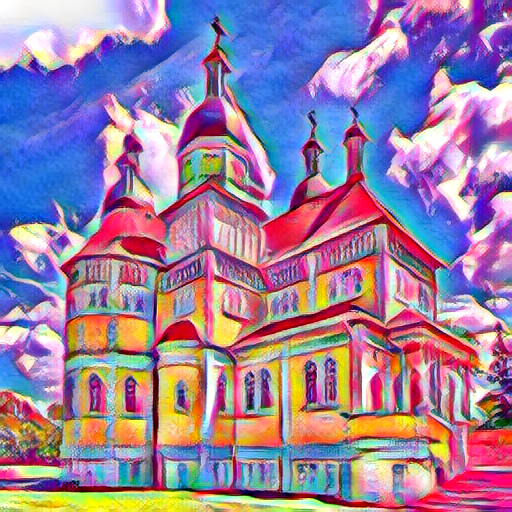}
    \end{minipage}
    \begin{minipage}{0.131\linewidth}
         \centering             \includegraphics[width=0.84\linewidth]{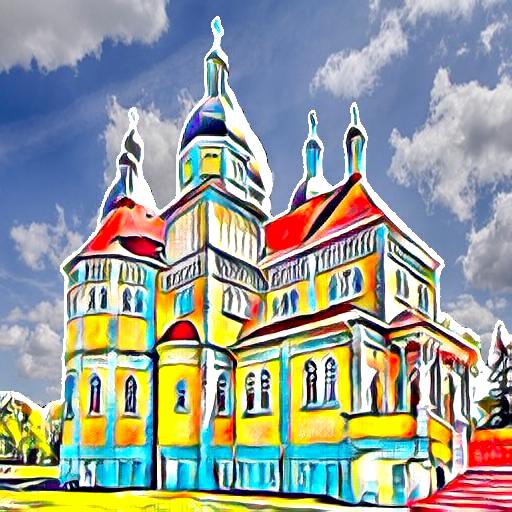}
    \end{minipage}    
    \caption{\footnotesize \textbf{Content Mismatch.} This figure illustrates content mismatch issues in style transfer methods, such as fire appearing on the car (first row) and graffiti style features on the building (second row). It can be observed that ObjMST (Ours) effectively minimizes these mismatches, producing more coherent stylized outputs.}
    \label{fig:method_CM}
\end{figure*}
\vspace{-0.4cm}
 \section{Experimental Results}
\label{sec:exp}
We evaluate our method across three tasks: 1) single-condition stylization through text; 2) single condition stylization through multimodal input on salient objects; 3) and double-condition stylization using text input for both salient and surrounding elements \footnote{Extended results are provided in the supplementary material.}
% We also conducted ablation studies to validate the efficacy of our method. 
\subsection{Text-Based IST (Single) \\~}
\label{subsec:TIST_Single}
\vspace{-0.3cm}
% we apply style-texts (column 1) as input on the facial images.
\noindent \textbf{Qualitative Analysis.}
As shown in Fig.~\ref{fig:outputs1}, CS \cite{kwon2022clipstyler} distorts the original content and compromises the structure of stylized output, while SemCS \cite{kamra2023sem} improves some outputs but still over-stylizes. CLVA \cite{fu2022language} poorly matches styles to text, often producing the same color. MMIST \cite{Wang2024WACV} applies styles across the entire image, neglecting salient content. In contrast, our method accurately stylizes only the foreground, preserving content without affecting the background. 

Fig.~\ref{fig:method_CM} illustrates content mismatch in the MMIST~\cite{Wang2024WACV}, CLVA~\cite{fu2022language}, SemCS~\cite{kamra2023sem}, and CS~\cite{kamra2023sem} methods, where style features undesirably affect the ground and sky in both the examples. In contrast, ObjMST confines style application to objects such as car and building, ensuring better semantic coherence. 

\begin{table}[]
% \begin{center}\renewcommand{\arraystretch}{1.00}
\scriptsize
\caption{\footnotesize \textbf{Style Representations Evaluation}. $T_S$, $I_S$, and $S$ represent the input style text, image, and generated style representations.  Each of these has corresponding foreground ($fg$) and background {$bg$} equivalents, i.e. $T_S$=$T_{S_{fg}}$and $T_{S_{bg}}$, $I_S$=$I_{S_{fg}}$and $I_{S_{bg}}$ and $S$=$S_{fg}$and $S_{bg}$. Here, $<>$ refers to cosine similarity. The average Clipscore~\cite{hessel2021clipscore} between generated style representations and image-text pair is higher, while the LPIPS~\cite{zhang2018unreasonable}  is lower with ObjMST.}
\footnotesize
\begin{tabular}{|l|ccl|l|}
\hline
\multirow{2}{*}{\textbf{Method}} & \multicolumn{3}{c|}{\textbf{Clipscore}$\uparrow$~\cite{hessel2021clipscore}}                                                                           & \multirow{2}{*}{\textbf{LPIPS}$\downarrow$}~\cite{zhang2018unreasonable} \\ \cline{2-4}
                                 & \multicolumn{1}{l|}{\textless{}$T_S$, S\textgreater{}} & \multicolumn{1}{l|}{\textless{}$I_S$,S\textgreater{}} & Average &                                 \\ \hline
\textbf{MMIST}~\cite{Wang2024WACV}                   & \multicolumn{1}{c|}{0.76}                          & \multicolumn{1}{c|}{0.48}                          & 0.46    & \multicolumn{1}{c|}{0.31}      \\ \hline
\textbf{Ours}                    & \multicolumn{1}{c|}{0.88}                          & \multicolumn{1}{c|}{0.56}                          & \textbf{0.51}    & \multicolumn{1}{c|}{\textbf{0.26}}      \\ \hline
\end{tabular}
\label{tab:sty_reps}
\end{table}
% ObjMST achieves higher Clipscore~\cite{hessel2021clipscore} than MMIST~\cite{Wang2024WACV}.
\noindent \textbf{Style Representations Quantitative Evaluation.} We evaluate generated style representations $S$, consisting of both salient and surrounding ($S_{fg}$  and $S_{bg}$) style representations using Clipscore~\cite{hessel2021clipscore} and LPIPS~\cite{zhang2018unreasonable} score. A higher Clipscore~\cite{hessel2021clipscore}, indicates better alignment of salient and surrounding style representations ($S_{fg}$ and $S_{bg}$) with respect to multimodal input style text-image pairs (\{$T_{S_{fg}}$, $I_{S_{fg}}$\} and \{$T_{S_{bg}}$, $I_{S_{bg}}$\}, respectively). This improved alignment occurs because the proposed loss function excludes irrelevant content features from CLIP embedding directions during the generation process. Consequently, the style representations are closely aligned with the style text (column 2) and style image (column 3) of Table~\ref{tab:sty_reps}. Similarly, a lower LPIPS~\cite{zhang2018unreasonable} score indicates better perceptual quality of generated style representations. We use foreground-input ($I_{S_{fg}}$) and background-input ($I_{S_{bg}}$) style image as reference image for computing LPIPS score on salient and surrounding style representations.  

% , because the masking operation used in the computation of foreground-input and salient-style CLIP embedding directions ($\Delta I_{fg}^k$ in Eq.~\ref{eq:eq1} and $\Delta S_{fg}^j$ in Eq.~\ref{eq:eq2}); background-input and surrounding-style CLIP embedding directions ($\Delta I_{bg}^k$ in Eq.~\ref{eq:eq4} and $\Delta S_{bg}^j$ in Eq.~\ref{eq:eq5}) ensures style alignment in salient and surrounding style representations.

 % Similarly, the style alignment is ensured in surrounding-style representations through Eq. ~\ref{eq:eq4} and \ref{eq:eq5}.
% in surrounding representations is ensured through the masking operation ($I_{C}^{'}$ and $M_C^{'}$) used in computing background-input and surrounding-style embedding directions $\Delta I_{bg}^k$ and $\Delta S_{bg}^j$ (Eq.~\ref{eq:eq4} and \ref{eq:eq5}).
% and background-input and surrounding style CLIP embedding directions ($\Delta I_{bg}$  in Eq.~\ref{eq:eq3} and $\Delta S_{bg}$ in Eq.~\ref{eq:eq4})

% the masking operation in proposed loss ensures that irrelevant background information does not dilute the style alignment, or introduce noise into perceptual calculations when computing LPIPS~\cite{zhang2018unreasonable} score.

% Style REps
% We report average Clipscore (column 4) and LPIPS~\cite{zhang2018unreasonable} of synthesized style representations (S) in column 4 of Table~\ref{tab:sty_reps}. Average Clipscore~\cite{hessel2021clipscore} is computed by taking the cosine similarity between intermediate style representations with respect to input style text, $T_S$ (column 2) and input style image, $I_S$ (column 3). 

\noindent \textbf{Stylized Outputs Quantitative Evaluation.} 
% For CS~\cite{kwon2022clipstyler}, SemCS~\cite{kamra2023sem}, and LDAST~\cite{fu2022language}, we computed the cosine similarity between the intermediate style representations (S) and the input style text ($T_S$). Clipscore~\cite{hessel2021clipscore} by taking cosine similarity between intermediate style representations (S) and  For MMIST~\cite{Wang2024WACV} and ObjMST, we computed this similarity on input style image ($I_S$). The average results of these methods are reported in column 4 of Table~\ref{tab:sty_reps}. 
We evaluated stylized outputs using both Full-Reference (FR) based -~ clipscore~\cite{hessel2021clipscore}, LPIPS~\cite{zhang2018unreasonable}, Contrique-FR~\cite{madhusudana2022image} and Non-Reference(NR) based -~ Nima~\cite{talebi2018nima} and Contrique (NR)~\cite{madhusudana2022image} metrics. For FR metrics, the input style image serves as the reference image for contrique-FR and LPIPS, while the input style text is considered as the reference text for clipscore~\cite{hessel2021clipscore} across all baseline methods. ObjMST outperforms baseline methods (See Table~\ref{table:scores}); demonstrating superior visual quality, measured by NIMA~\cite{talebi2018nima}); enhanced perceptual quality, evaluated by LPIPS~\cite{zhang2018unreasonable}, Contrique~\cite{madhusudana2022image}) in both FR and NR evaluations; and better alignment of style text with the stylized outputs, measured by clipscore~\cite{hessel2021clipscore}. Additionally, we conducted a user study to support this evaluation.\\~
\begin{table}[]
% \begin{center}\renewcommand{\arraystretch}{0.25}
% \small
\caption{\footnotesize \textbf{Stylized Outputs Evaluation, TIST (Single).} \footnotesize The table shows that quantitative scores (Nima~\cite{talebi2018nima}, Contrique~\cite{madhusudana2022image} (FR, NR)), Clipscore~\cite{hessel2021clipscore} and LPIPS~\cite{zhang2018unreasonable} obtained with our method (ObjMST) outperform those of the baseline methods. \footnotesize}
% \small
\begin{tabular}{llllll}
\hline
\textbf{Scores}                                                        & \multicolumn{1}{c}{\begin{tabular}[c]{@{}c@{}}CS\\ \scriptsize~\cite{kwon2022clipstyler}\end{tabular}} & \multicolumn{1}{c}{\begin{tabular}[c]{@{}c@{}}SemCS\\ \scriptsize~\cite{kamra2023sem}\end{tabular}} & \multicolumn{1}{c}{\begin{tabular}[c]{@{}c@{}}LDAST\\ \scriptsize~\cite{fu2022language}\end{tabular}} & \multicolumn{1}{c}{\begin{tabular}[c]{@{}c@{}}MMIST\\ \scriptsize~\cite{Wang2024WACV}\end{tabular}} & \multicolumn{1}{c}{\begin{tabular}[c]{@{}c@{}}Ours\\ \end{tabular}} \\ \hline
\textbf{Nima}$\uparrow$\scriptsize~\cite{talebi2018nima}                                                          & 5.05                                                              & 4.81                                                                 & 4.90                                                                 & 4.75                                                                 & \textbf{5.14}                                                       \\ \hline
\textbf{\begin{tabular}[c]{@{}l@{}}Contrique\\ \scriptsize(FR)$\uparrow$~\cite{madhusudana2022image}\end{tabular}}  & 0.28                                                              & 0.24                                                                 & 0.30                                                                 & 0.29                                                                 & \textbf{0.32}                                                       \\ \hline
\textbf{\begin{tabular}[c]{@{}l@{}}Contrique\\\scriptsize(NR)$\uparrow$~\cite{madhusudana2022image}\end{tabular}} & 33.36                                                             & 29.88                                                               & 43.36                                                                & 47.56                                                                & \textbf{48.14}                                                      \\ \hline
\textbf{Clipscore}$\uparrow$\scriptsize~\cite{hessel2021clipscore}                                                     & 0.38                                                               & 0.40                                                                  & 0.41                                                                  & \multicolumn{1}{c}{0.42}                                                 & \multicolumn{1}{c}{\textbf{0.46}}                                                \\ \hline
\textbf{LPIPS}$\downarrow$\scriptsize~\cite{zhang2018unreasonable}                                                       & 0.45                                                               & 0.37                                                                  & 0.35                                                                  & \multicolumn{1}{c}{0.40}                                                 & \multicolumn{1}{c}{\textbf{0.32}}                                                \\ \hline
\textbf{User Study}$\uparrow$                                                   & 18.98                                                             & 14.92                                                                & 14.15                                                                & 23.89                                                                & \textbf{28.06}                                                      \\ \hline
\end{tabular}
\label{table:scores}
\end{table}
\noindent \textbf{\textit{User Study}.} We applied 20 multimodal styles to 30 distinct content images, generating 600 stylized images per method. One hundred participants evaluated the images in a user study, voting for those that demonstrated the best style consistency and content preservation, particularly in the foreground of the content image. Table~\ref{table:scores} presents the percentage of votes, highlighting the superior performance of the proposed ObjMST.

\begin{figure*}[!htb]
       \centering
   \begin{minipage}{0.090\linewidth}
         % \centering
         % \raggeedleft
        \tiny         
            \textbf{Style Text /} \\
            \tiny
            \textbf{Content Image}
        \end{minipage}
        \hspace{0.1cm}
     \begin{minipage}{0.080\linewidth}
         % \centering
         \footnotesize
        \textbf{F:} Copper plate engraving. \\
        \textbf{B:} Underwater
    \end{minipage}
    \hspace{-0.3cm}
     \begin{minipage}{0.048\linewidth}
         \centering             \includegraphics[width=0.60cm, height=0.60cm]{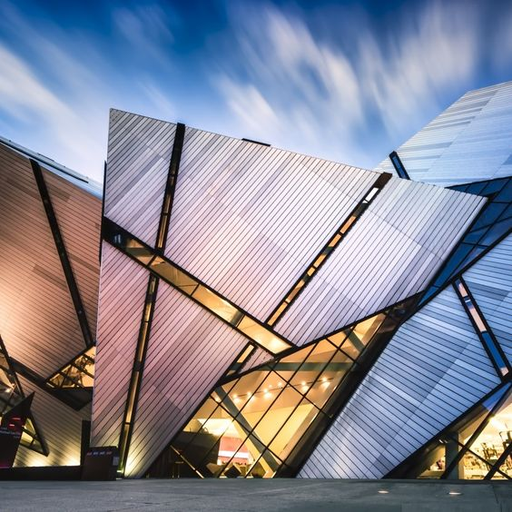}
        \end{minipage}
        \hspace{0.2cm}
     \begin{minipage}{0.080\linewidth}
         % \centering
         \scriptsize
        \textbf{F:} A fauvism style painting with bright colour. \\
        \textbf{B:} Seascape
    \end{minipage}
    \hspace{-0.3cm}
      \begin{minipage}{0.048\linewidth}
         \centering             \includegraphics[width=0.60cm, height=0.60cm]{images/0467_resized.png}
        \end{minipage}
        \hspace{0.05cm}
    \begin{minipage}{0.080\linewidth}
         % \centering
         \scriptsize
         % by Katsushika Hokusai.
        \textbf{F}: The great wave off King Kanagawa \\
        \textbf{B}: Glitter
    \end{minipage}
    \hspace{-0.3cm}
     \begin{minipage}{0.048\linewidth}
         \centering             \includegraphics[width=0.60cm, height=0.60cm]{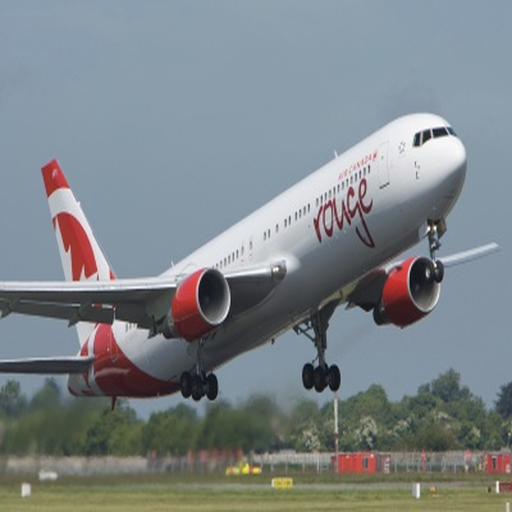}
        \end{minipage}
        \hspace{0.05cm}
     \begin{minipage}{0.080\linewidth}
         % \centering
         \scriptsize
        \textbf{F:} Fantasy Vivid Colors. \\
        \textbf{B:} Starry Night.
    \end{minipage}
    \hspace{-0.3cm}
    \begin{minipage}{0.048\linewidth}
         \centering             \includegraphics[width=0.60cm, height=0.60cm]{images/0976_resized.png}
        \end{minipage}
        \hspace{0.1cm}
     \begin{minipage}{0.080\linewidth}
         % \centering
        \scriptsize
        \textbf{F:} Lisa Frank. \\
        \textbf{B:} Ice.
    \end{minipage}
    \hspace{-0.3cm}
    \begin{minipage}{0.048\linewidth}
         \centering             \includegraphics[width=0.60cm, height=0.60cm]{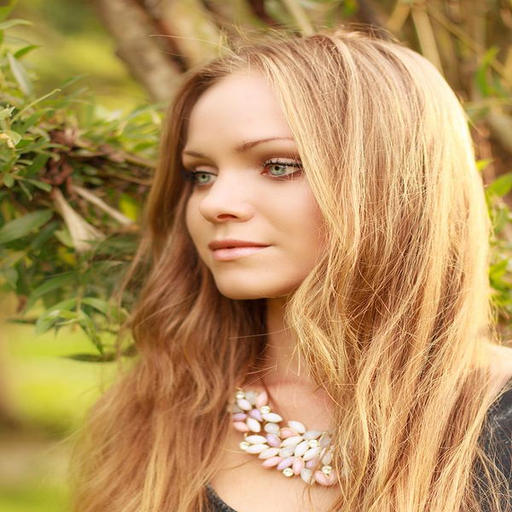}
        \end{minipage}
        \hspace{0.2cm}
     \begin{minipage}{0.080\linewidth}
         % \centering
         \scriptsize
        \textbf{F:} Green Crystal. \\
        \textbf{B:} The Great Wave off Kanagawa.
    \end{minipage}  
    \hspace{-0.3cm}
    \begin{minipage}{0.048\linewidth}
         \centering
         \includegraphics[width=0.60cm, height=0.60cm]{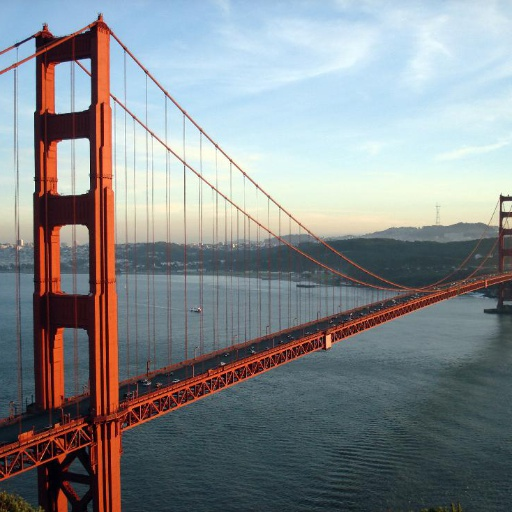}
        \end{minipage}
      \begin{minipage}{0.100\linewidth}
         \centering
         \scriptsize
            SemCS~\cite{kamra2023sem}
        \end{minipage}
            \begin{minipage}{0.138\linewidth}
         \centering
             \includegraphics[width=0.83\linewidth]{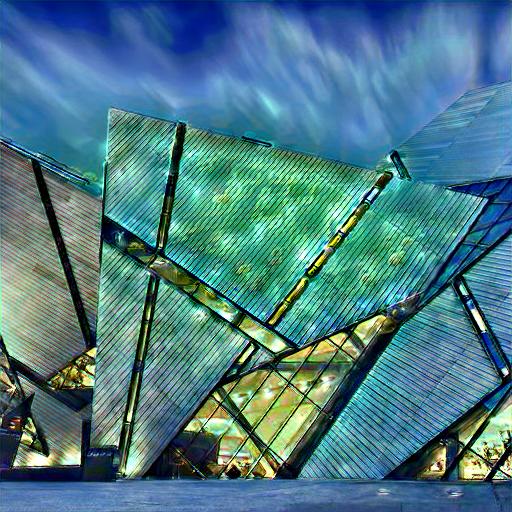}
        \end{minipage}
        \begin{minipage}{0.138\linewidth}
         \centering             \includegraphics[width=0.83\linewidth]{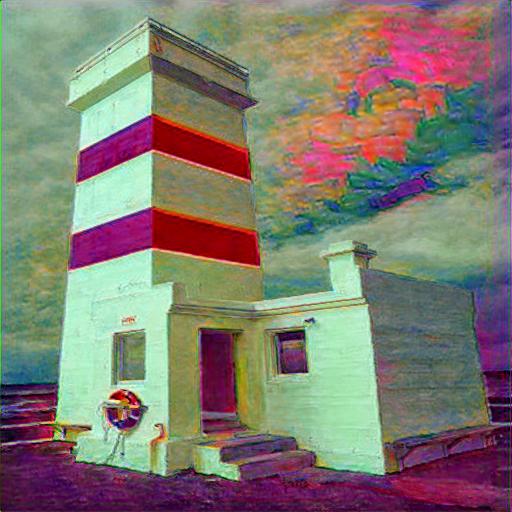}
        \end{minipage}
    \begin{minipage}{0.138\linewidth}
         \centering
             \includegraphics[width=0.83\linewidth]{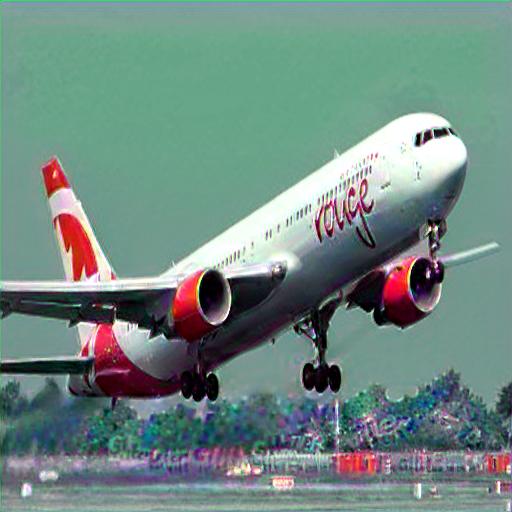}
        \end{minipage}
        \begin{minipage}{0.138\linewidth}
         \centering
             \includegraphics[width=0.83\linewidth]{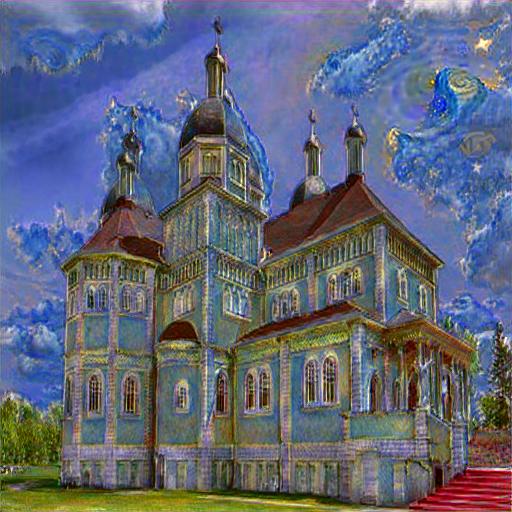}
        \end{minipage}
    \begin{minipage}{0.138\linewidth}
         \centering
             \includegraphics[width=0.83\linewidth]{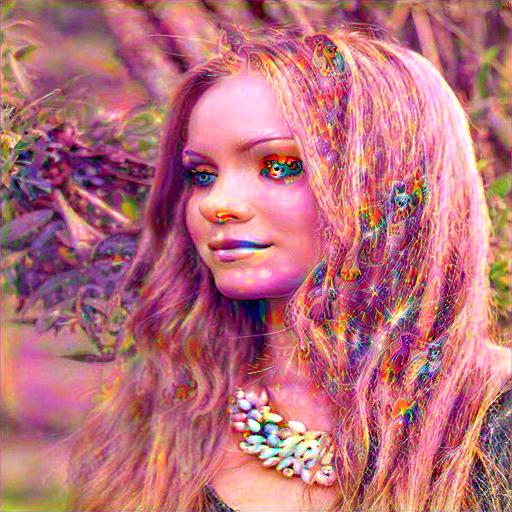}
        \end{minipage}
    \begin{minipage}{0.138\linewidth}
         \centering
             \includegraphics[width=0.83\linewidth]{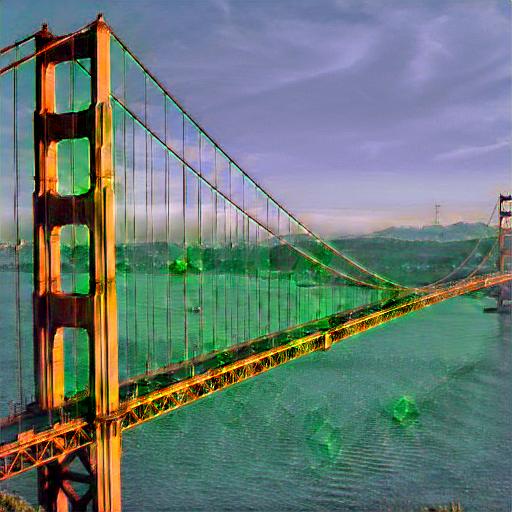}             
        \end{minipage}
        \begin{minipage}{0.100\linewidth}
         \centering
         \scriptsize
            \textbf{Ours}
        \end{minipage}        
        % \hfill
        \begin{minipage}{0.138\linewidth}
         \centering
             \includegraphics[width=0.83\linewidth]{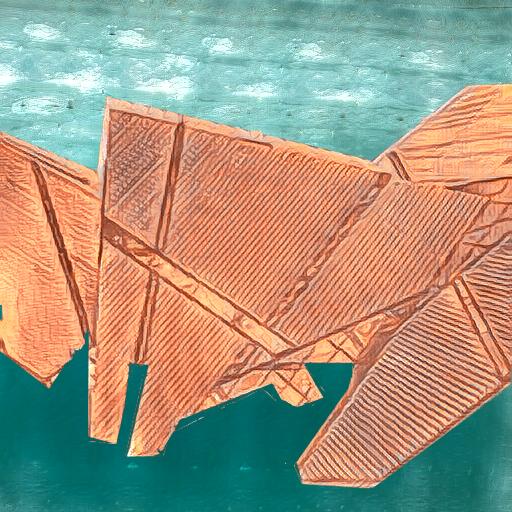}
        \end{minipage}
        \begin{minipage}{0.138\linewidth}
         \centering
             \includegraphics[width=0.83\linewidth]{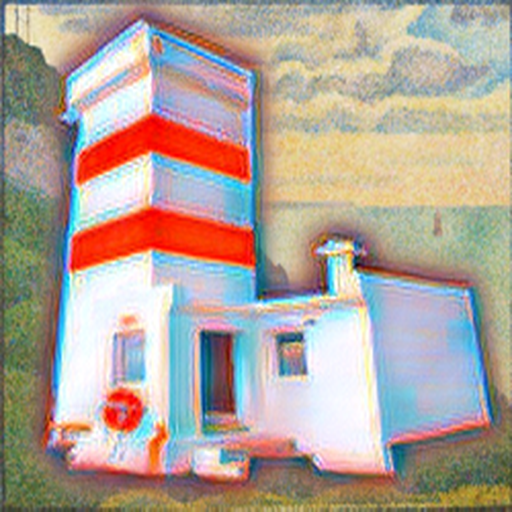}
        \end{minipage}
        \begin{minipage}{0.138\linewidth}
         \centering
             \includegraphics[width=0.83\linewidth]{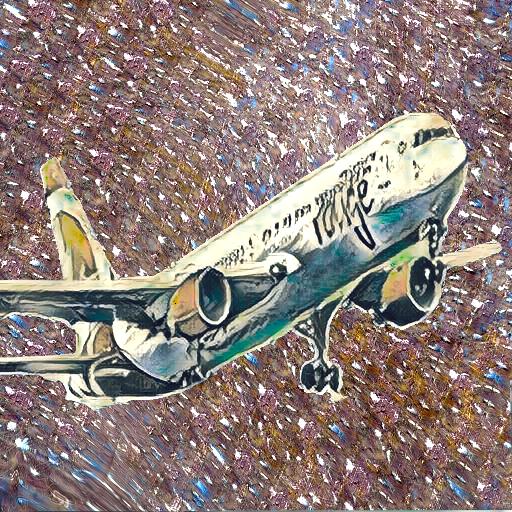}
        \end{minipage}
        \begin{minipage}{0.138\linewidth}
         \centering
             \includegraphics[width=0.83\linewidth]{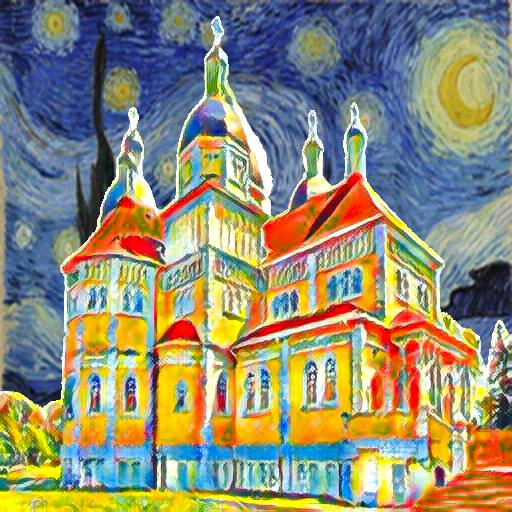}
        \end{minipage}
        \begin{minipage}{0.138\linewidth}
         \centering
             \includegraphics[width=0.83\linewidth]{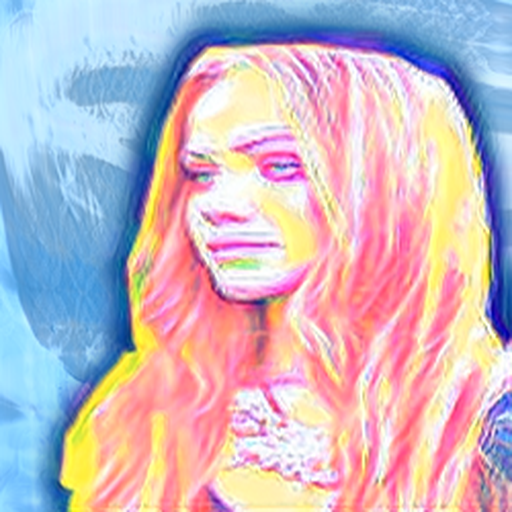}
        \end{minipage}
        \begin{minipage}{0.138\linewidth}
         \centering
             \includegraphics[width=0.83\linewidth]{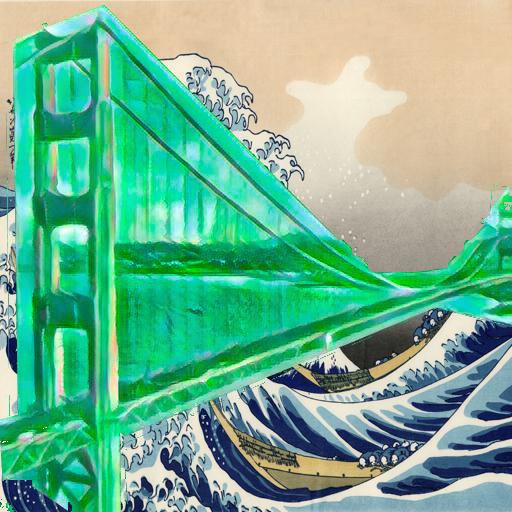}
        \end{minipage}   
    % \vspace{-0.1cm}
        \caption{\footnotesize \textbf{Text-based IST (double).} Distinct text style features are applied to the foreground (\textbf{F}) and the background (\textbf{B}). In SemCS [17], the absence of a clear boundary between salient objects and surrounding elements results in subpar quality of stylized outputs. In contrast, our method produces outputs with a distinct separation between foreground and background style features. Furthermore, our method provides a superior representation of style features compared to SemCS \cite{kamra2023sem}.}
        \label{fig:double_style_res}
    \end{figure*}
\subsection{Text-Based IST (Double)}  
Fig.~\ref{fig:double_style_res} demonstrates that our method consistently applies distinct style representations to the foreground and background based on the provided style text, avoiding the entanglement of salient and surrounding elements, unlike SemCS~\cite{kamra2023sem}. For instance, in column 1, our approach applies copper plate style features to the salient object and "underwater" features to the surrounding elements. In contrast, SemCS~\cite{kamra2023sem} fails to maintain a clear boundary, incorrectly stylizing foreground and background.
\subsection{Multimodal IST (Single)} 
\begin{figure}[!htb]
       \centering
   \begin{minipage}{0.100\linewidth}
         % \centering
         \scriptsize
            \textbf{Style Text}
        \end{minipage}
    % \begin{minipage}{0.138\linewidth}
    %      % \centering
    %      \scriptsize
    %     A Cubism Style
    % \end{minipage}
     \begin{minipage}{0.188\linewidth}
         % \centering
         \scriptsize
        The Great Wave off Kanagawa.       
    \end{minipage}
     \begin{minipage}{0.188\linewidth}
         \centering
         \scriptsize
       Impressionism         
    \end{minipage}
    %  \begin{minipage}{0.080\linewidth}
    %      \centering
    %      \small
    %     Ice        
    % \end{minipage}
     \begin{minipage}{0.188\linewidth}
         \centering
         \scriptsize
        A Baroque painting.
    \end{minipage}
     \begin{minipage}{0.188\linewidth}
         \centering
         \scriptsize
        Lisa Frank
    \end{minipage} 
   \begin{minipage}{0.100\linewidth}
         \centering
         \scriptsize
            \textbf{Style Image}
        \end{minipage}
    % \begin{minipage}{0.138\linewidth}
    %      \centering
    %          \includegraphics[width=0.99\linewidth]{images/Cubism.jpeg}
    %     \end{minipage}
     \begin{minipage}{0.188\linewidth}
         \centering
             \includegraphics[width=0.49\linewidth]{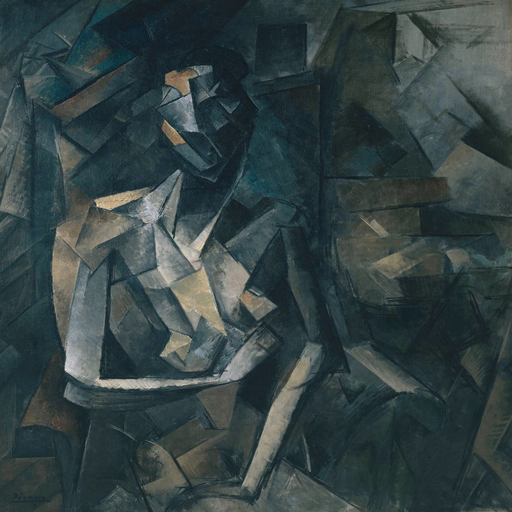}
        \end{minipage}
    \begin{minipage}{0.188\linewidth}
         \centering
             \includegraphics[width=0.49\linewidth]{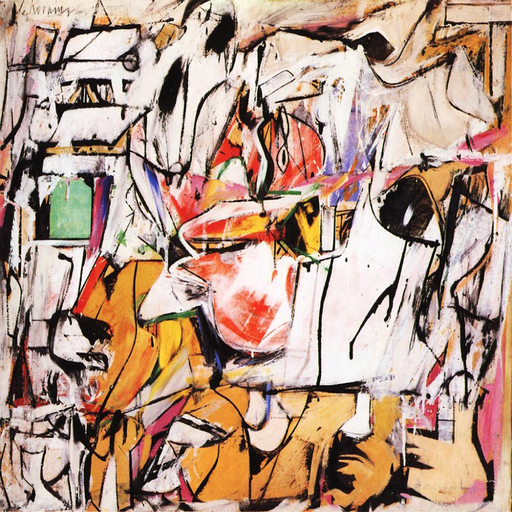}
        \end{minipage} 
      % \begin{minipage}{0.138\linewidth}
      %    \centering
      %        \includegraphics[width=0.99\linewidth]{images/trial.png}
      %   \end{minipage}
    \begin{minipage}{0.188\linewidth}
         \centering
             \includegraphics[width=0.49\linewidth]{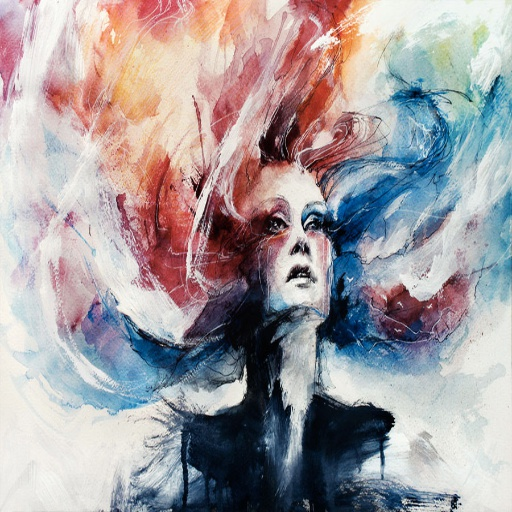}
        \end{minipage}
    \begin{minipage}{0.188\linewidth}
         \centering
             \includegraphics[width=0.49\linewidth]{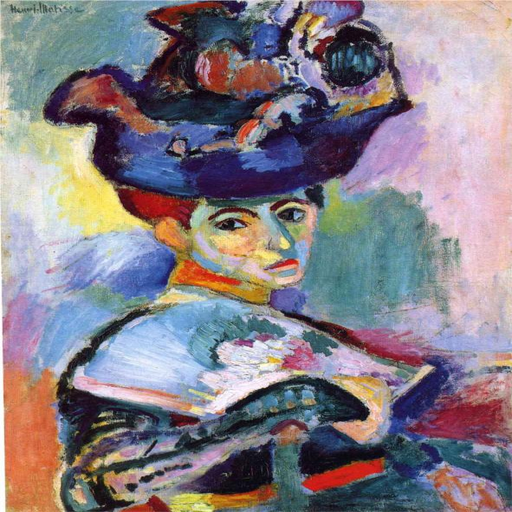}
        \end{minipage}
      \begin{minipage}{0.108\linewidth}
         \centering
         \small
            MMIST~\cite{Wang2024WACV}
        \end{minipage}
    % \begin{minipage}{0.138\linewidth}
    %      \centering   
    %        \includegraphics[width=0.99\linewidth]{images/Acubismstylepainting_0958_resized_mmist.jpg}
    %     \end{minipage}
            \begin{minipage}{0.188\linewidth}
         \centering
             \includegraphics[width=0.99\linewidth]{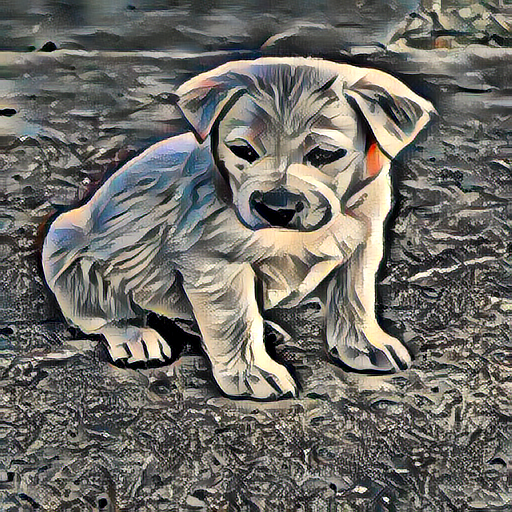}
        \end{minipage}
        \begin{minipage}{0.188\linewidth}
         \centering
             \includegraphics[width=0.99\linewidth]{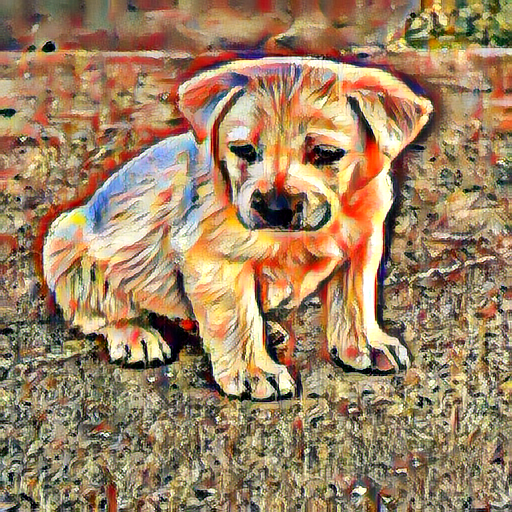}
        \end{minipage}
    % \begin{minipage}{0.138\linewidth}
    %      \centering
    %          \includegraphics[width=0.99\linewidth]{images/Ice_0958_resized_mmist.jpg}
    %     \end{minipage}
    \begin{minipage}{0.188\linewidth}
         \centering
             \includegraphics[width=0.99\linewidth]{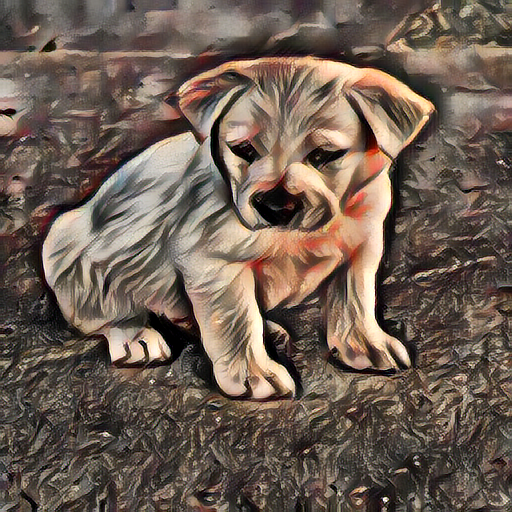}
    \end{minipage}
   \begin{minipage}{0.188\linewidth}
         \centering
             \includegraphics[width=0.99\linewidth]{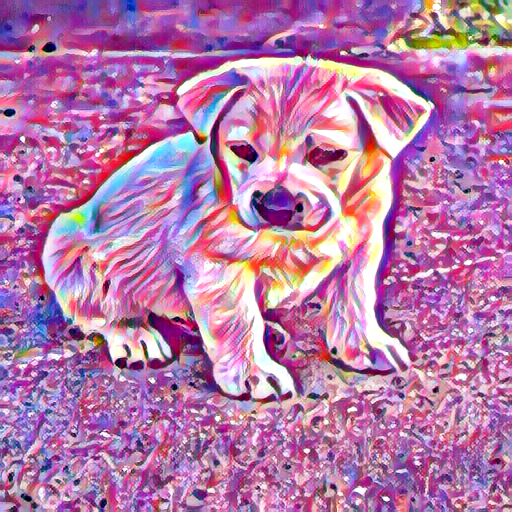}
    \end{minipage}
     \begin{minipage}{0.108\linewidth}
         \centering
         \small
            Ours
        \end{minipage}        
        % \hfill
        % \begin{minipage}{0.138\linewidth}
        %  \centering
        %      \includegraphics[width=0.99\linewidth]{images/Acubismstylepainting_0958_resized_ours.jpg}
        % \end{minipage}
        \begin{minipage}{0.188\linewidth}
         \centering
             \includegraphics[width=0.99\linewidth]{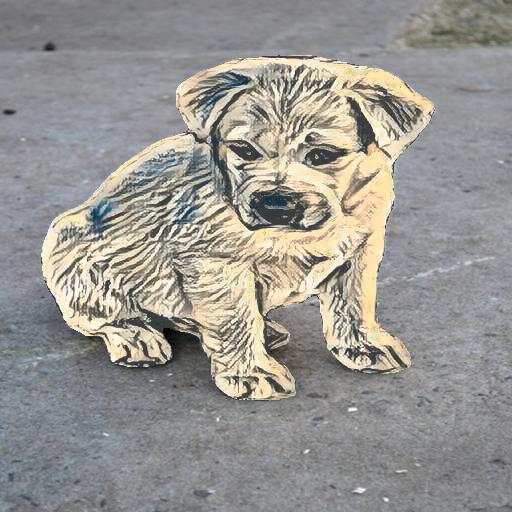}
        \end{minipage}
        \begin{minipage}{0.188\linewidth}
         \centering
             \includegraphics[width=0.99\linewidth]{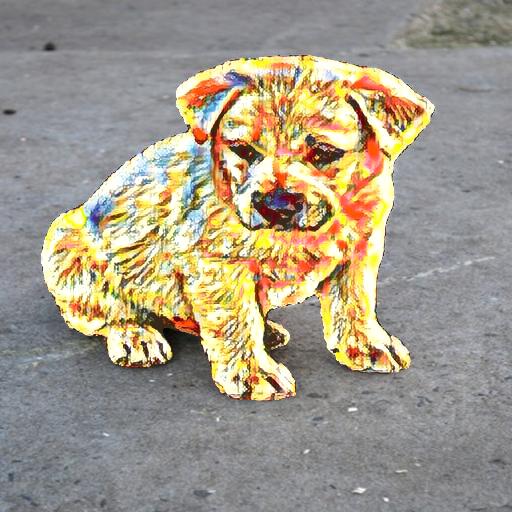}
        \end{minipage}
        % \begin{minipage}{0.138\linewidth}
        %  \centering
        %      \includegraphics[width=0.99\linewidth]{images/Ice_0958_resized_ours.jpg}
        % \end{minipage}
        \begin{minipage}{0.188\linewidth}
         \centering
             \includegraphics[width=0.99\linewidth]{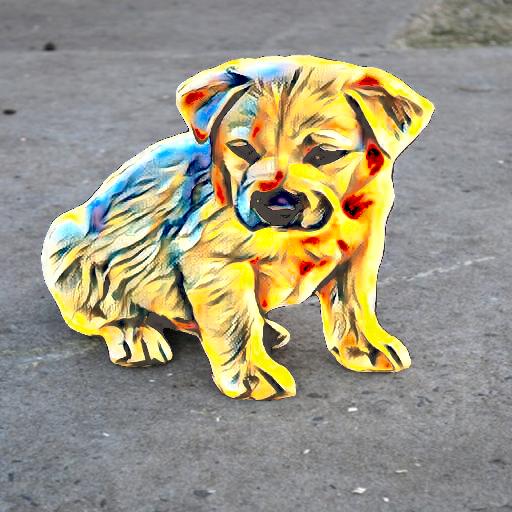}
        \end{minipage}
        \begin{minipage}{0.188\linewidth}
         \centering
             \includegraphics[width=0.99\linewidth]{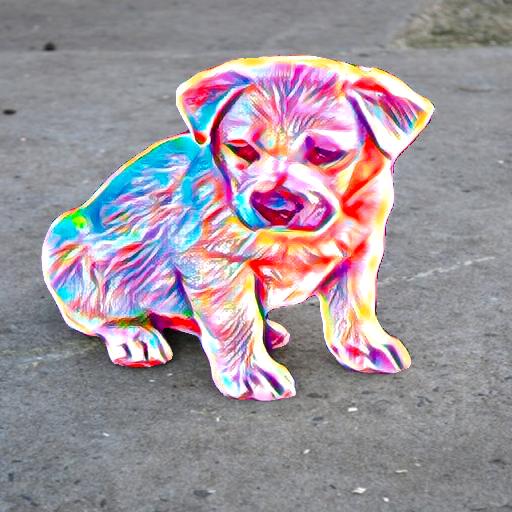}
        \end{minipage}   
        \caption{\footnotesize \textbf{Multimodal IST (single).} MMIST (third row) applies style to the entire image, while ObjMST (fourth row) applies it only to the object.}    
        \label{fig:multimodal}
    \end{figure}

As shown in Fig.~\ref{fig:multimodal}, MMIST (third row) distributes the style over the entire image, affecting both the foreground and background elements. In contrast, ObjMST (fourth row) selectively applies the style only to the object, leaving the surrounding areas unaffected. Unlike MMIST \cite{Wang2024WACV}, which applies style uniformly using the All-to-All attention mechanism, ObjMST employs Salient-to-Key (S2K) attention mechanism to apply style only to salient objects(only Puppy’s face and body).
\subsection{Ablation Studies}
\label{subsec:ablation}
\begin{figure}[!htb]
       \centering
           \begin{minipage}{0.119\linewidth}
     \centering     
        \hspace{0.001cm} 
    \end{minipage}
    \begin{minipage}{0.119\linewidth}
     \centering
     \small
        i)
    \end{minipage}
    \begin{minipage}{0.119\linewidth}
     \centering         ii)
    \end{minipage}
    \begin{minipage}{0.119\linewidth}
     \centering
            iii)
    \end{minipage}
    \begin{minipage}{0.119\linewidth}
     \centering
     iv)
    \end{minipage}
    \begin{minipage}{0.119\linewidth}
     \centering
     v)
    \end{minipage}    
    \begin{minipage}{0.119\linewidth}
     \centering
     vi)
    \end{minipage} 
    % \begin{minipage}{0.119\linewidth}
    %  \centering
    %  vii)
    % \end{minipage} 
             \begin{minipage}{0.119\linewidth}
         \centering
         % \small
         \scriptsize
            w/o proposed Loss \textbf{(0.52)} 
        \end{minipage}
        %   \begin{minipage}{0.119\linewidth}
        %  \centering             \includegraphics[width=0.99\linewidth]{images/Ice0_MMIST.png}
        % \end{minipage}
            \begin{minipage}{0.119\linewidth}
         \centering             \includegraphics[width=0.99\linewidth]{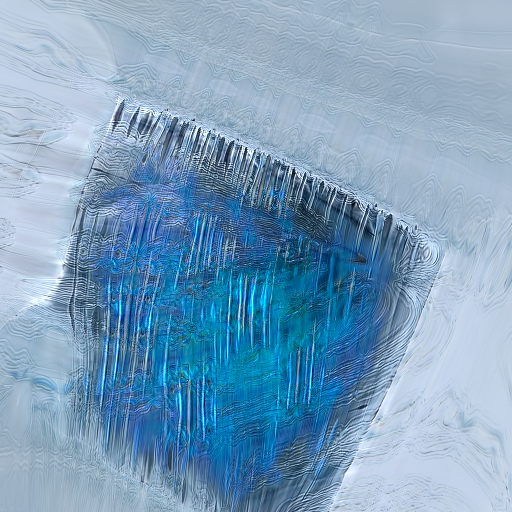}
        \end{minipage}
        \begin{minipage}{0.119\linewidth}
         \centering             \includegraphics[width=0.99\linewidth]{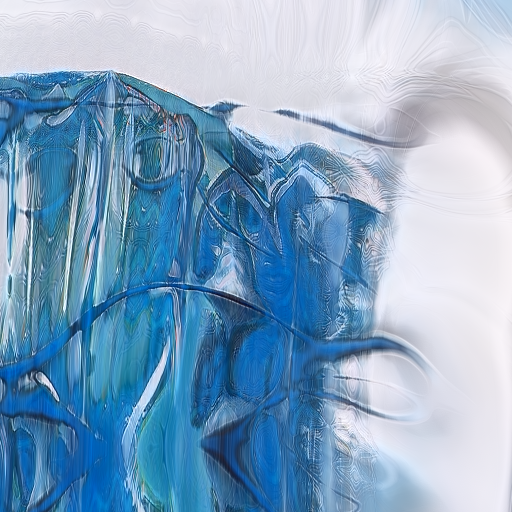}
        \end{minipage}
        \begin{minipage}{0.119\linewidth}
         \centering             \includegraphics[width=0.99\linewidth]{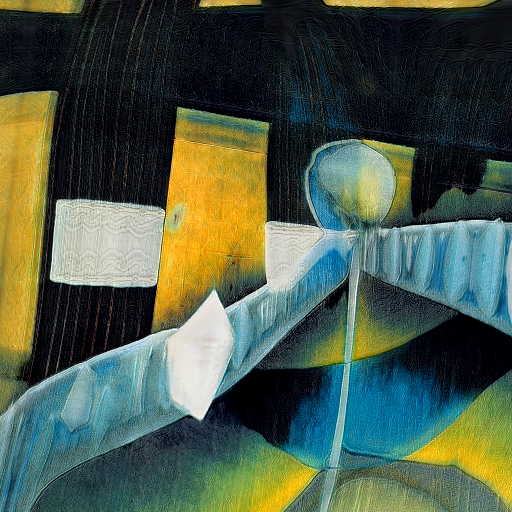}
        \end{minipage}
        \begin{minipage}{0.119\linewidth}
         \centering             \includegraphics[width=0.99\linewidth]{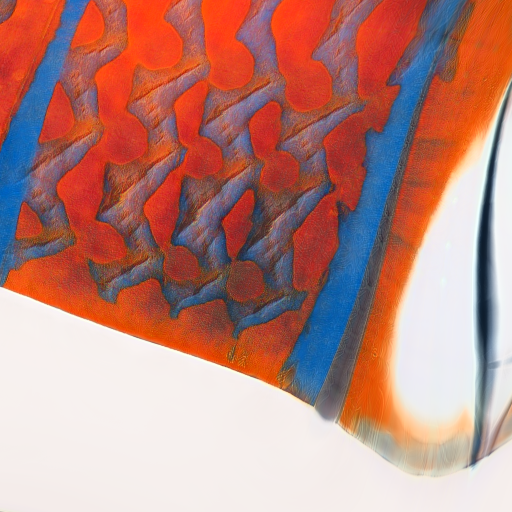}
        \end{minipage}
        % \begin{minipage}{0.119\linewidth}
        %  \centering
        %      \includegraphics[width=0.99\linewidth]{images/Ice5_MMIST.png}
        % \end{minipage}
        \begin{minipage}{0.119\linewidth}
         \centering             \includegraphics[width=0.99\linewidth]{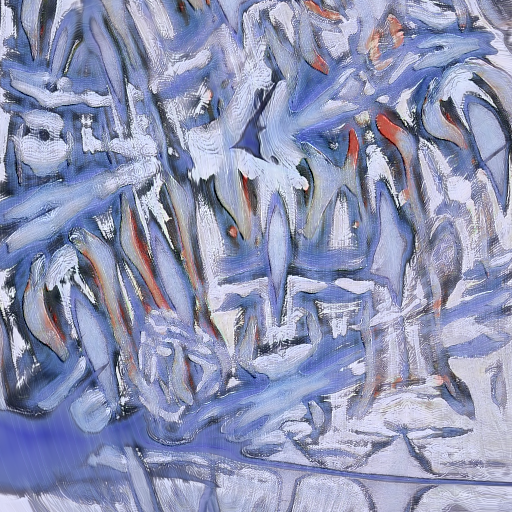}
        \end{minipage}
        \begin{minipage}{0.119\linewidth}
         \centering             \includegraphics[width=0.99\linewidth]{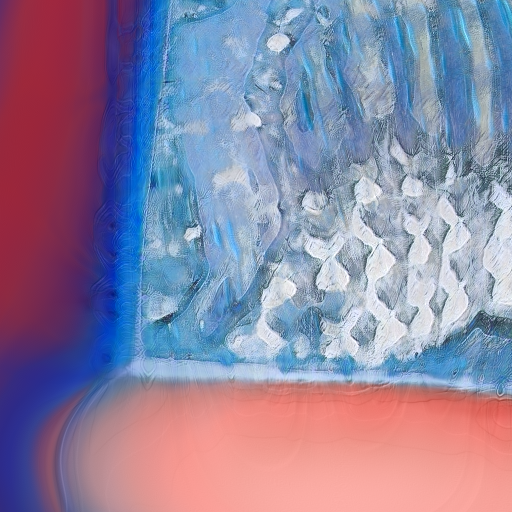}
        \end{minipage}
         \begin{minipage}{0.119\linewidth}
         \centering
         % \small
         \scriptsize
            with proposed Loss \textbf{(0.76)}
        \end{minipage}
          \begin{minipage}{0.119\linewidth}
         \centering             \includegraphics[width=0.99\linewidth]{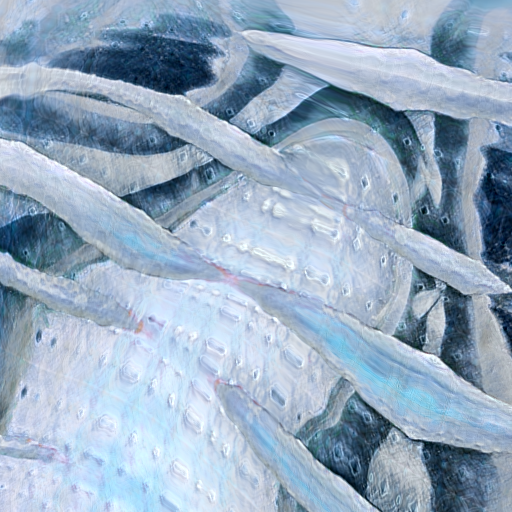}
        \end{minipage}
            \begin{minipage}{0.119\linewidth}
         \centering             \includegraphics[width=0.99\linewidth]{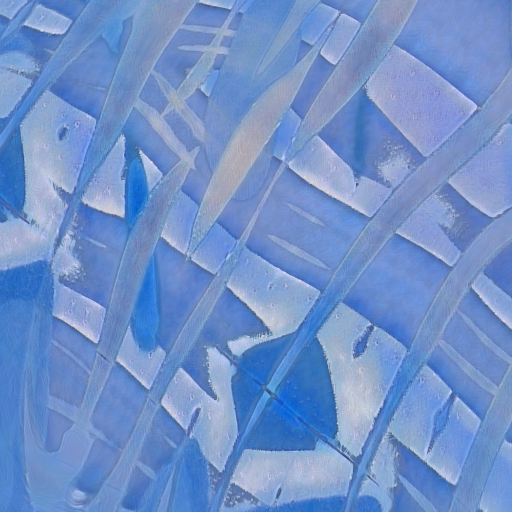}
        \end{minipage}
        % \begin{minipage}{0.119\linewidth}
        %  \centering             \includegraphics[width=0.99\linewidth]{images/Ice2_ours.png}
        % \end{minipage}
        \begin{minipage}{0.119\linewidth}
         \centering             \includegraphics[width=0.99\linewidth]{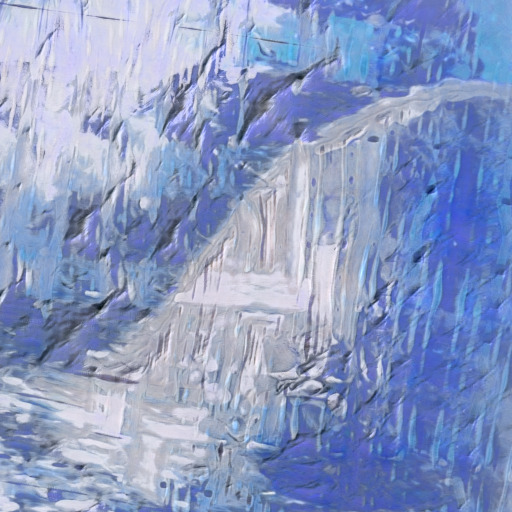}
        \end{minipage}
        \begin{minipage}{0.119\linewidth}
         \centering             \includegraphics[width=0.99\linewidth]{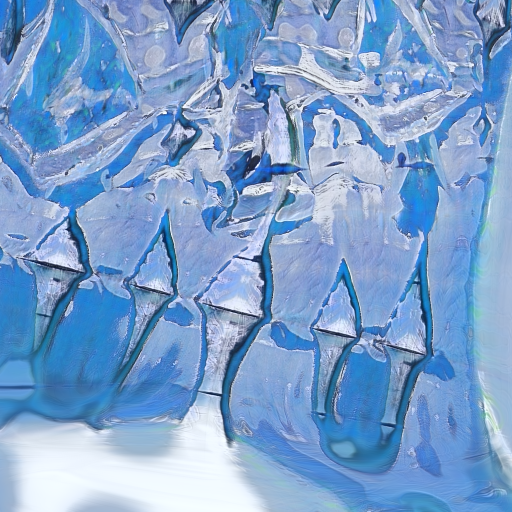}
        \end{minipage}
        \begin{minipage}{0.119\linewidth}
         \centering             \includegraphics[width=0.99\linewidth]{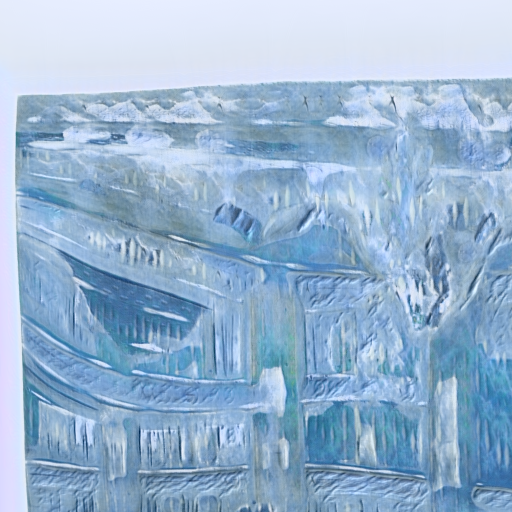}
        \end{minipage}
        \begin{minipage}{0.119\linewidth}
         \centering             \includegraphics[width=0.99\linewidth]{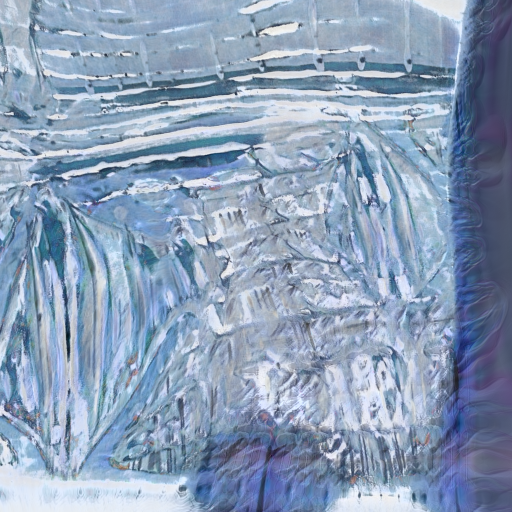}
        \end{minipage}
\caption{\footnotesize \textbf{Ablation Study I} Style representations (columns i-vi) generated with Masked Directional CLIP loss consistently align with the text "Ice." Misalignment occurs (columns iii) and iv) without this loss, resulting in a lower CLIP score (0.52).}
\label{fig:ablation1}
\end{figure}

\begin{figure}[!htb]
       \centering
    \begin{minipage}{0.108\linewidth}
         \centering
         \small
            \textbf {A2K}
    \end{minipage}
        \begin{minipage}{0.248\linewidth}
         \centering             \includegraphics[width=0.79\linewidth]{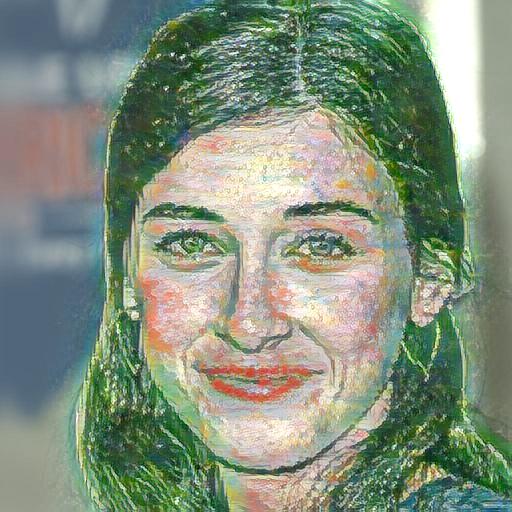}
        \end{minipage}
       \begin{minipage}{0.248\linewidth}
         \centering             \includegraphics[width=0.79\linewidth]{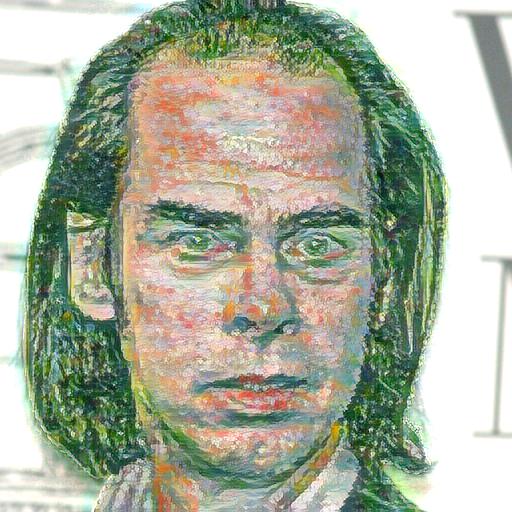}
        \end{minipage}
        \begin{minipage}{0.248\linewidth}
         \centering             \includegraphics[width=0.79\linewidth]{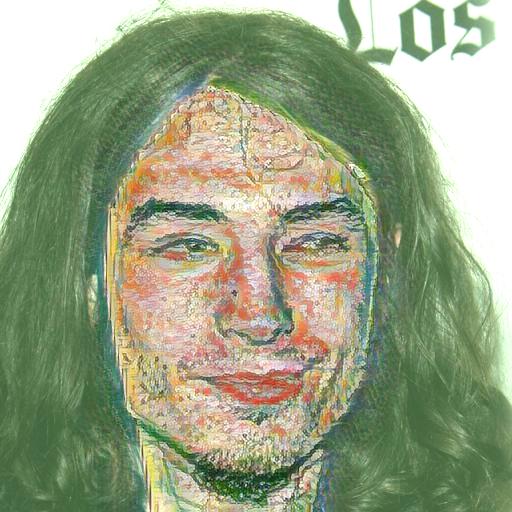}
        \end{minipage}
    \begin{minipage}{0.108\linewidth}
         \centering
         \small
            \textbf {S2K}
    \end{minipage}
        \begin{minipage}{0.248\linewidth}
         \centering             \includegraphics[width=0.79\linewidth]{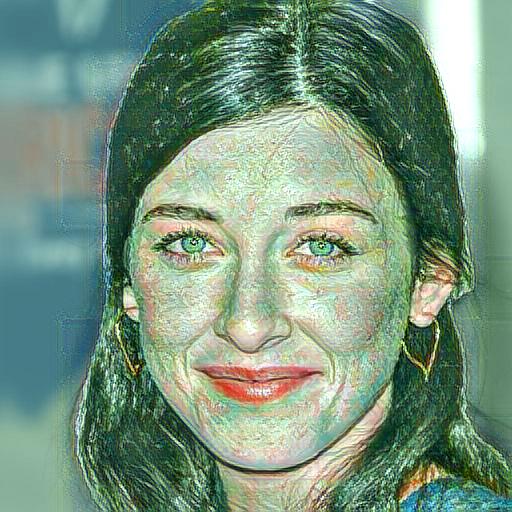}
        \end{minipage} 
        \begin{minipage}{0.248\linewidth}
         \centering             \includegraphics[width=0.79\linewidth]{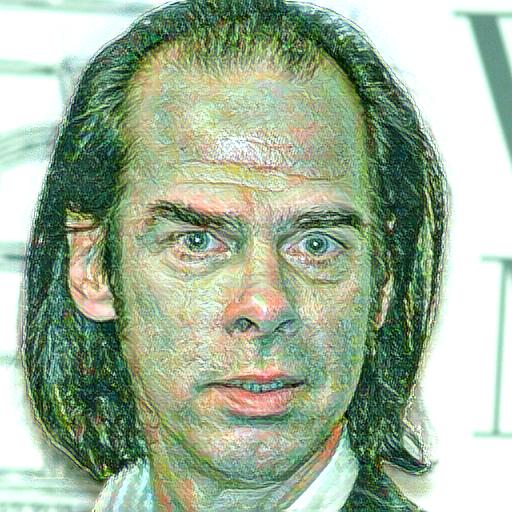}
        \end{minipage}
        \begin{minipage}{0.248\linewidth}
         \centering             \includegraphics[width=0.79\linewidth]{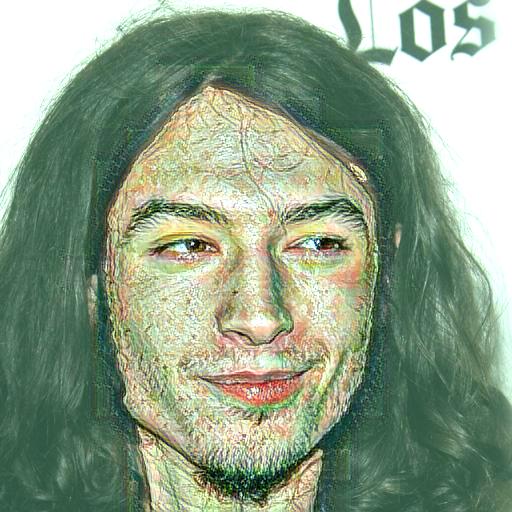}
        \end{minipage}
\caption{\footnotesize \textbf{Ablation Study II.} This study examines the effects of Salient-to-Key (S2K) mapping and All-to-All Key (A2K) mapping using the style input "A money style painting.}
\label{fig:ablation2}
\end{figure}
\vspace{-0.05cm}
\noindent \textbf{Ablation Study I.} 
This study examines multimodal style representations with (bottom row) and without (top row) proposed loss as shown in columns (i-vi) of Fig.~\ref{fig:ablation1}. MMIST\cite{Wang2024WACV} fails to capture "Ice" features (columns iii and iv), whereas our method (bottom row) consistently aligns style features with "Ice." This is because the proposed loss eliminates the noisy features. A higher Clipscore \cite{hessel2021clipscore} with the proposed loss function validates the consistency and alignment of the style representations with multimodal input.\\~

\noindent \textbf{Ablation Study II.}
Fig.~\ref{fig:ablation2} demonstrates that All-to-All Key (A2K) (top row) blurs the facial features on stylization, while Salient-to-Key (S2K) applies the style features without blurring the facial features and expressions. This difference arises due to the mapping from content to stylized outputs.

\vspace{-0.3cm}
\section{Conclusion}
In this work, we introduced ObjMST, an object-focused multimodal style transfer framework that addresses misalignment and content mismatch issues using a masked directional CLIP loss, Salient-To-Key (S2K) attention mechanism for stylizing salient objects, and image harmonization seamlessly blending the stylized objects with their surroundings. Our results demonstrate the framework's effectiveness in multimodal IST and text-based IST under both single and double-text conditions. As future work, we propose exploring multimodal IST for images with multiple objects, each stylized with distinct and unique representations.
\vspace{-0.3cm}
\bibliographystyle{cas-model2-names}
% \bibliographystyle{IEEEtran}

% \bibliography{prl2024_review}
% \bibliography{ref}

\newpage

\begin{center}
    {\Large Supplementary}
    % \Large ObjMST: Object-Focused Multimodal Style Transfer
\end{center}

 \begin{figure*}
    \centering
    \includegraphics[width=0.89\linewidth]{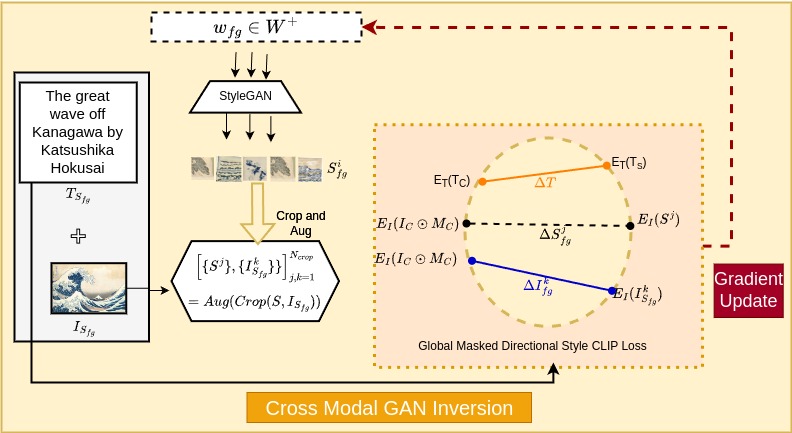}
    \vspace{0.2cm}
     \caption{Cross Modal GAN Inversion for multimodal input on the salient object.}
    \label{fig:crossmodal}
\end{figure*}

\section{Detailed Experimental Results} 
\noindent \textbf{Evaluation Metrics.} In Sec.~5.2 of the manuscript, we reported average scores of below evaluation metrics: \\~
\textit{NIMA \cite{talebi2018nima}:} This score predicts the distribution of human opinions about an image without a reference image. \\~
\textit{Contrique \cite{madhusudana2022image}:} This metric refers to the CONTRastive Image QUality Evaluator and aims to learn image quality representation in a self-supervised manner. It provides scores in two settings under Full-Reference as well as Non-Refernce.  \\~
We computed these average scores on 88 randomly sampled stylized obtained with different methods CS \cite{kwon2022clipstyler}, SemCS \cite{kamra2023sem}, LDAST \cite{fu2022language}, MMIST \cite{Wang2024WACV} and Ours. We presented detailed scores for each evaluation metric in \textcolor{blue}{Table 1-3} presented below. The average scores reported in the manuscript show that our method outperforms than baselines for applying style features through single text conditions on salient objects.\\~

\begin{figure*}
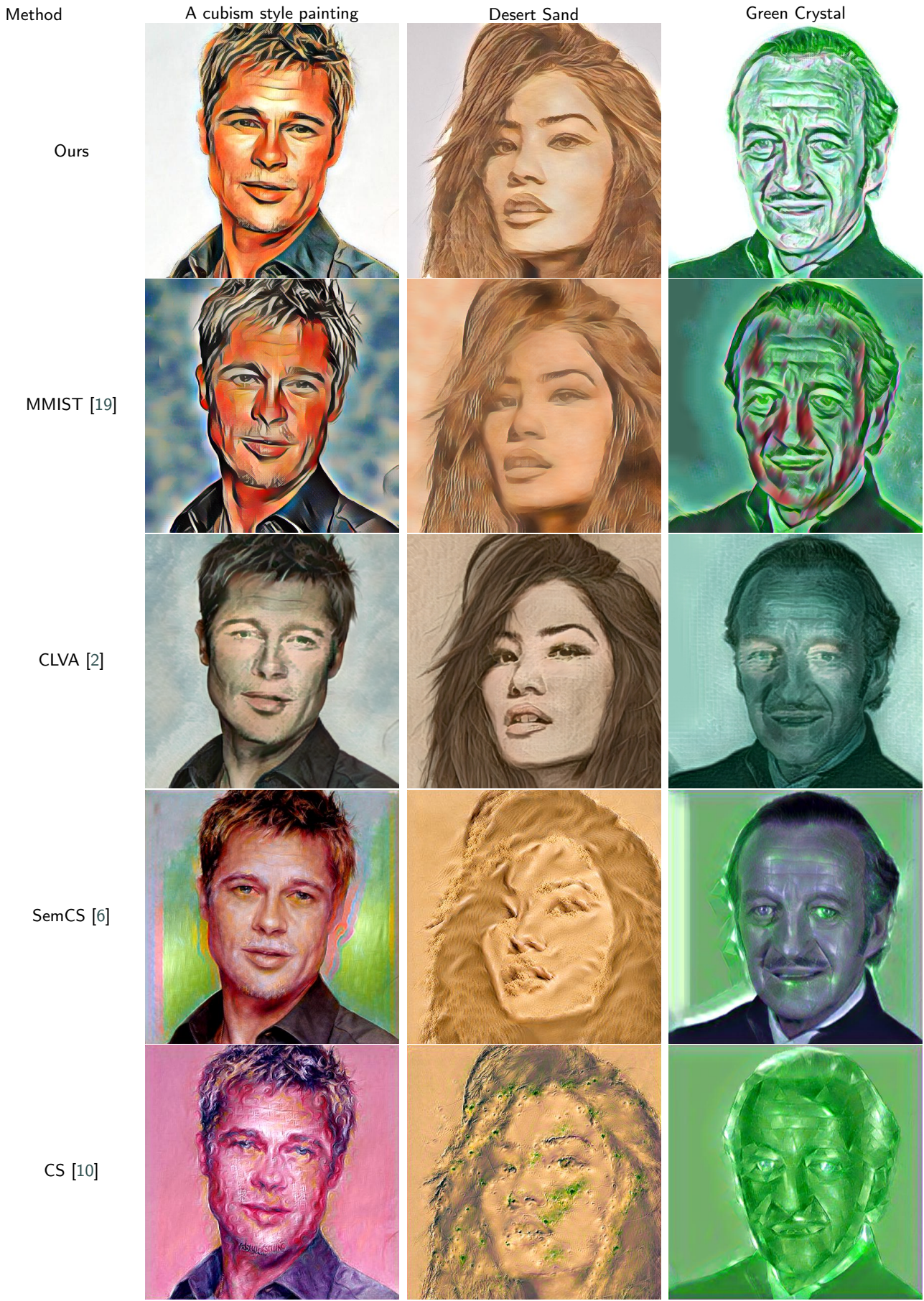

    \centering
   \begin{minipage}{0.14\linewidth}
     % \footnotesize  
     Method
    \end{minipage}
    \begin{minipage}{0.270\linewidth}
     \centering
     \begin{center}
      A cubism style painting   
     \end{center}      
    \end{minipage}    
   \begin{minipage}{0.270\linewidth}
     \centering
     % \scriptsize
     Desert Sand
    \end{minipage}
   \begin{minipage}{0.270\linewidth}
     \centering
     % \scriptsize
     Green Crystal
    \end{minipage}
    \begin{minipage}{0.14\linewidth}
     \centering
     % \footnotesize
     Ours
    \end{minipage}
   \begin{minipage}{0.270\textwidth}
         \centering             \includegraphics[width=0.99\linewidth]{images/Acubismstylepainting_brad_pitt_ours.jpg}
    \end{minipage}
  \begin{minipage}{0.270\textwidth}
         \centering            
         \includegraphics[width=0.99\linewidth]{images/Desertsand_25_ours.jpg}
    \end{minipage}
  \begin{minipage}{0.270\textwidth}
         \centering        \includegraphics[width=0.99\linewidth]{images/GreenCrystal_37_ours.jpg}
    \end{minipage}

    \begin{minipage}{0.14\linewidth}
     \centering
     % \footnotesize
     MMIST \cite{Wang2024WACV}
    \end{minipage}
   \begin{minipage}{0.270\textwidth}
         \centering             \includegraphics[width=0.99\linewidth]{images/Acubismstylepainting_brad_pitt_mmist.jpg}
    \end{minipage}
  \begin{minipage}{0.270\textwidth}
         \centering            
         \includegraphics[width=0.99\linewidth]{images/Desertsand_25_mmist.jpg}
    \end{minipage}
  \begin{minipage}{0.270\textwidth}
         \centering            
         \includegraphics[width=0.99\linewidth]{images/GreenCrystal_37_mmist.jpg}
    \end{minipage}
    \begin{minipage}{0.14\linewidth}
     \centering
     % \footnotesize
     CLVA \cite{fu2022language}
    \end{minipage}
   \begin{minipage}{0.270\textwidth}
         \centering             \includegraphics[width=0.99\linewidth]{images/Acubismstylepainting_brad_pitt_ldast.png}
    \end{minipage}
  \begin{minipage}{0.270\textwidth}
         \centering            
         \includegraphics[width=0.99\linewidth]{images/DesertSand_25_ldast.jpg}
    \end{minipage}
  \begin{minipage}{0.270\textwidth}
         \centering            
         \includegraphics[width=0.99\linewidth]{images/GreenCrystal_37_ldast.jpg}
    \end{minipage}
   \begin{minipage}{0.14\linewidth}
     \centering
     % \footnotesize
     SemCS \cite{kamra2023sem}
    \end{minipage}
   \begin{minipage}{0.270\textwidth}
         \centering             \includegraphics[width=0.99\linewidth]{images/SemCS_Acubismstylepainting_brad_pitt_brad_pitt.jpg}
    \end{minipage}
  \begin{minipage}{0.270\textwidth}
         \centering            
         \includegraphics[width=0.99\linewidth]{images/SemCS_Sand_25_25.jpg}
    \end{minipage}
  \begin{minipage}{0.270\textwidth}
         \centering            
         \includegraphics[width=0.99\linewidth]{images/SemCS_GreenCrystal_37_37.jpg}
    \end{minipage}
   \begin{minipage}{0.14\linewidth}
     \centering
     % \scriptsize
     CS \cite{kwon2022clipstyler}
    \end{minipage}
   \begin{minipage}{0.270\textwidth}
         \centering             \includegraphics[width=0.99\linewidth]{images/CS_Acubismstylepainting_brad_pitt_BradPitt.jpg}
    \end{minipage}
   \begin{minipage}{0.270\textwidth}
         \centering            
         \includegraphics[width=0.99\linewidth]{images/CS_DesertSand_25_25.jpg}
    \end{minipage}
   \begin{minipage}{0.270\textwidth}
         \centering            
         \includegraphics[width=0.99\linewidth]{images/CS_GreenCrystal_37_37.jpg}
    \end{minipage}    
    \caption{\textbf{TIST (Single)}. Text conditions on top of each column are the style text applied to the content image.}
    \label{fig: fig1}
  \end{figure*}

\begin{figure*}
    \centering
   \begin{minipage}{0.14\linewidth}
     % \footnotesize  
     Method
    \end{minipage}
    \begin{minipage}{0.270\linewidth}
     \centering
     \begin{center}
     Fire   
     \end{center}      
    \end{minipage}    
   \begin{minipage}{0.270\linewidth}
     \centering
     % \scriptsize
     A graffiti Style Painting
    \end{minipage}
   \begin{minipage}{0.270\linewidth}
     \centering
     % \scriptsize
     Ice
    \end{minipage} 
    \begin{minipage}{0.14\linewidth}
     \centering
     % \footnotesize
     Ours
    \end{minipage}
  \begin{minipage}{0.270\textwidth}
         \centering            
         \includegraphics[width=0.99\linewidth]{images/fire_0592_resized_ours.jpg}
    \end{minipage} 
  \begin{minipage}{0.270\textwidth}
         \centering            
         \includegraphics[width=0.99\linewidth]{images/Agraffitistylepainting_0976_resized_ours.jpg}
    \end{minipage} 
  \begin{minipage}{0.270\textwidth}
         \centering            
         \includegraphics[width=0.99\linewidth]{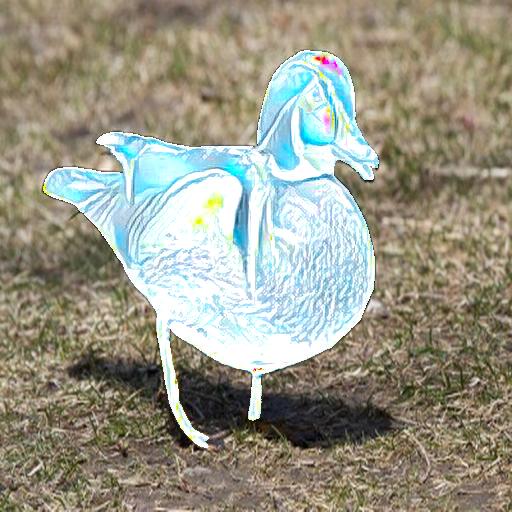}
    \end{minipage} 
    \begin{minipage}{0.14\linewidth}
     \centering
     % \footnotesize
     MMIST \cite{Wang2024WACV}
    \end{minipage}
  \begin{minipage}{0.270\textwidth}
         \centering            
         \includegraphics[width=0.99\linewidth]{images/Fire_0592_resized_mmist.jpg}
    \end{minipage}   
  \begin{minipage}{0.270\textwidth}
         \centering            
         \includegraphics[width=0.99\linewidth]{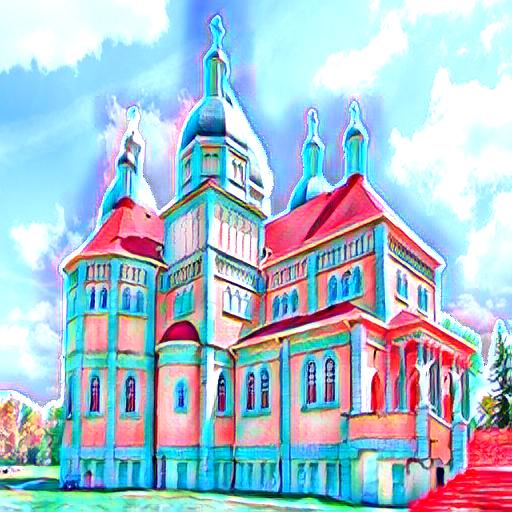}
    \end{minipage} 
  \begin{minipage}{0.270\textwidth}
         \centering            
         \includegraphics[width=0.99\linewidth]{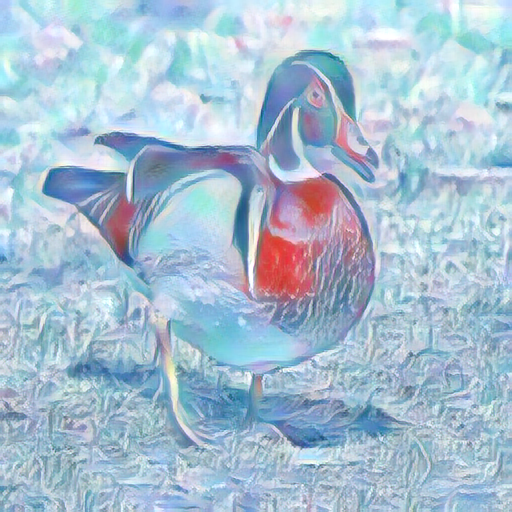}
    \end{minipage} 
     \begin{minipage}{0.14\linewidth}
     \centering
     % \footnotesize
     LDAST \cite{fu2022language}
    \end{minipage}   
      \begin{minipage}{0.270\textwidth}
         \centering            
         \includegraphics[width=0.99\linewidth]{images/Fire_0592_resized_ldast.png}
    \end{minipage}
          \begin{minipage}{0.270\textwidth}
         \centering            
         \includegraphics[width=0.99\linewidth]{images/AgraffitiStylePainting_0976_resized_Ldast.png}
    \end{minipage}  
  \begin{minipage}{0.270\textwidth}
         \centering            
         \includegraphics[width=0.99\linewidth]{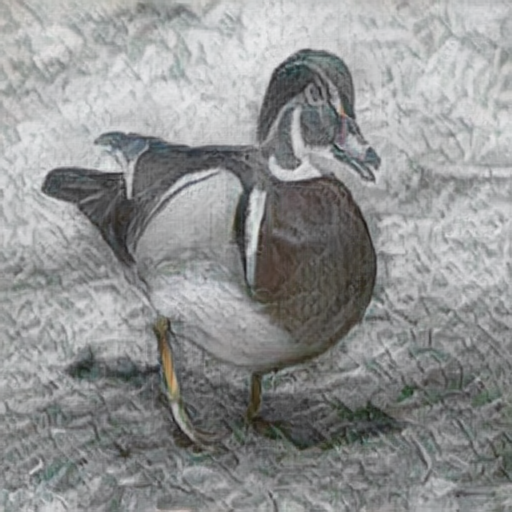}
    \end{minipage} 
     \begin{minipage}{0.14\linewidth}
     \centering
     % \footnotesize
     SemCS \cite{kamra2023sem}
    \end{minipage} 
 \begin{minipage}{0.270\textwidth}
         \centering            
         \includegraphics[width=0.99\linewidth]{images/SemCS_Fire_0592_resized_0592_resized.jpg}
    \end{minipage}
          \begin{minipage}{0.270\textwidth}
         \centering            
         \includegraphics[width=0.99\linewidth]{images/SemCS_Agraffitistylepainting_0976_resized_0976_resized.jpg}
    \end{minipage}  
  \begin{minipage}{0.270\textwidth}
         \centering            
         \includegraphics[width=0.99\linewidth]{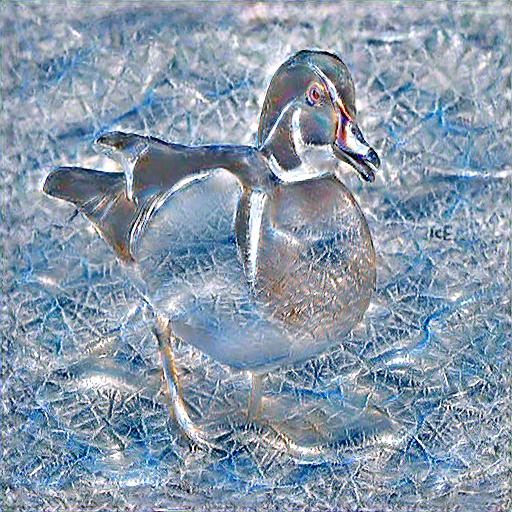}
    \end{minipage} 
     \begin{minipage}{0.14\linewidth}
     \centering
     % \footnotesize
     CS \cite{kamra2023sem}
     \end{minipage}
    \begin{minipage}{0.270\textwidth}
         \centering            
         \includegraphics[width=0.99\linewidth]{images/CS_Fire_0592_resized_0592_resized.jpg}
    \end{minipage}
    \begin{minipage}{0.270\textwidth}
         \centering            
         \includegraphics[width=0.99\linewidth]{images/CS_Agraffitistylepainitng_0976_resized_0976_resized.jpg}
    \end{minipage}
  \begin{minipage}{0.270\textwidth}
         \centering            
         \includegraphics[width=0.99\linewidth]{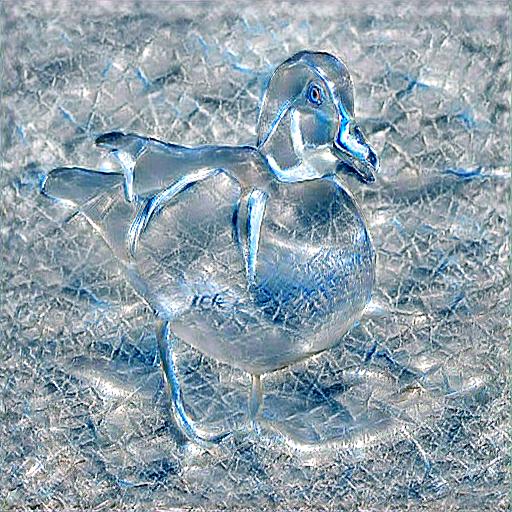}
    \end{minipage} 
    
    \caption{\textbf{TIST (Single)}. Text conditions on top of each column are the style text applied to the content image.}
    \label{fig: fig2}
\end{figure*}

\noindent \textbf{Text-Guided IST Outputs (Single).} Fig. \ref{fig: fig1} and \ref{fig: fig2} show the stylized outputs obtained with our proposed and SemCS \cite{kamra2023sem} methods using double Text Condition. These outputs are the extended view of Figures 1 and 4 of the main manuscript. Clearly, stylized outputs obtained with our proposed method show superior visual quality as compared to baseline methods. \\~

\begin{figure*}
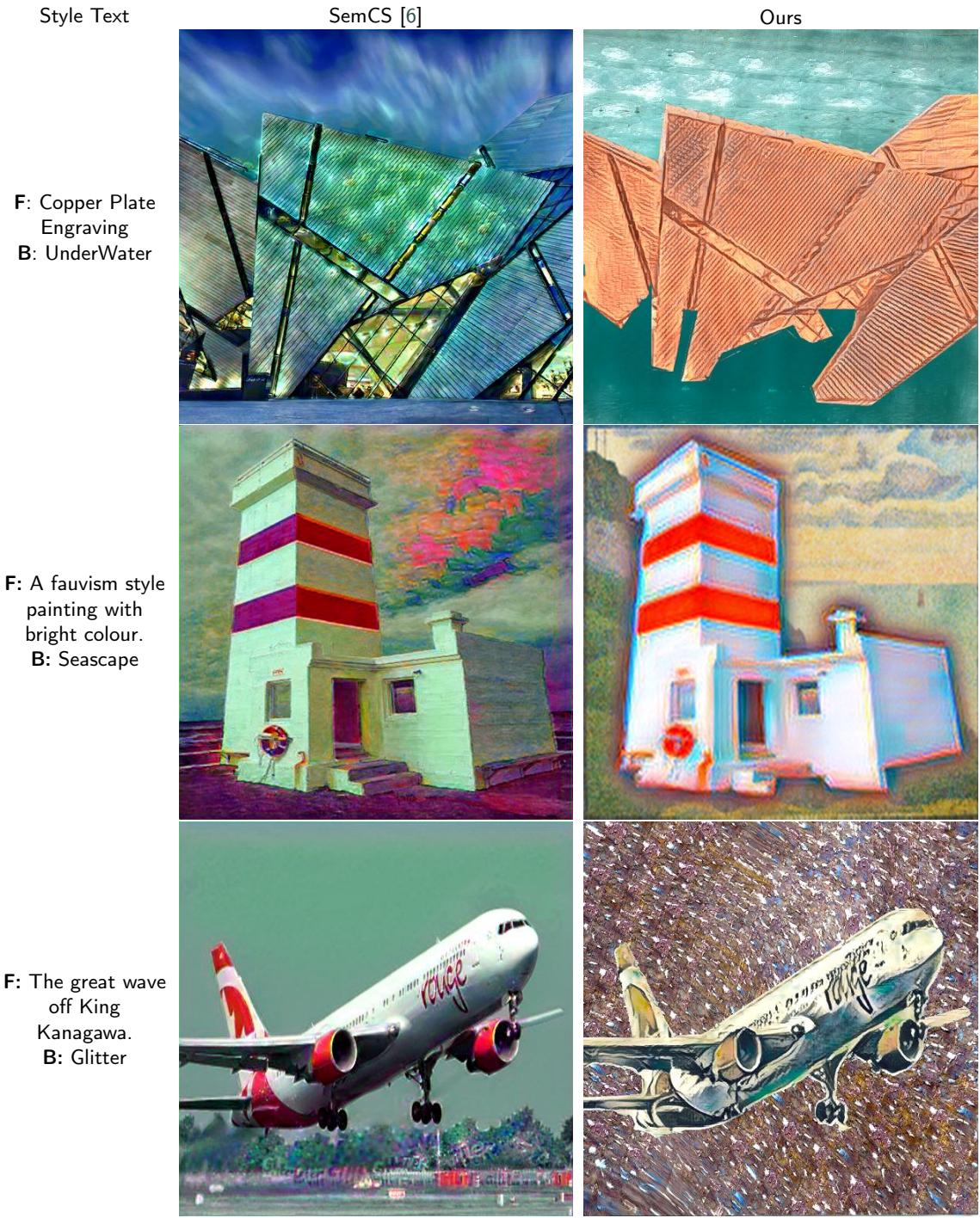

    \centering
          \begin{minipage}{0.15\linewidth}
       \centering
            Style Text 
        \end{minipage}
    \begin{minipage}{0.35\linewidth}
    \centering
            SemCS \cite{kamra2023sem}
        \end{minipage}
    \begin{minipage}{0.35\linewidth}
    \centering
            Ours
        \end{minipage}      
    \begin{minipage}{0.15\linewidth}         
    \centering
        \textbf{F}: Copper Plate Engraving \\
        \textbf{B}: UnderWater
        \end{minipage}         
    \begin{minipage}{0.35\linewidth}
         \centering             \includegraphics[width=0.99\linewidth]{images/SemCSglobfb_Copperplateengraving_underwater_modern.jpg}
        \end{minipage}
    \begin{minipage}{0.35\linewidth}
         \centering             \includegraphics[width=0.99\linewidth]{images/Copperplateengraving+copper_modern_ours.jpg}
        \end{minipage}
    \begin{minipage}{0.15\linewidth}         
    \centering
        \textbf{F:} A fauvism style painting with bright colour. \\
        \textbf{B:} Seascape
        \end{minipage}         
    \begin{minipage}{0.35\linewidth}
         \centering             \includegraphics[width=0.99\linewidth]{images/SemCSglobfb_Seascape_Afauvismstylepainting_0467_resized.jpg}
        \end{minipage}
    \begin{minipage}{0.35\linewidth}
         \centering             \includegraphics[width=0.99\linewidth]{images/seascape_Afauvismstylepaintingwithbrightcolor_0467_resized.png}
        \end{minipage}
 \begin{minipage}{0.15\linewidth}         
    \centering
        \textbf{F:} The great wave off King Kanagawa. \\
        \textbf{B:} Glitter
        \end{minipage}         
    \begin{minipage}{0.35\linewidth}
         \centering             \includegraphics[width=0.99\linewidth]{images/SemCSglobfb_TheGreatWaveoffKanagawa_Glitter_0961_resized.jpg}
        \end{minipage}
    \begin{minipage}{0.35\linewidth}
         \centering             \includegraphics[width=0.99\linewidth]{images/ThegreatwaveoffKanagawaHoksuai_0961_resized_ours.jpg}
        \end{minipage}
    \caption{\textbf{TIST (Double).} We apply separate style feature on salient object and surrounding elements of content image.}
    \label{fig:fig3}
\end{figure*}

\begin{figure*}
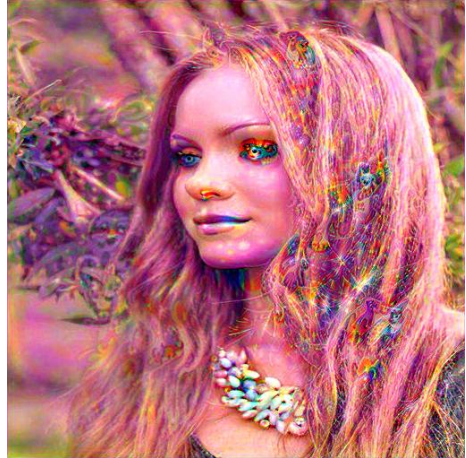
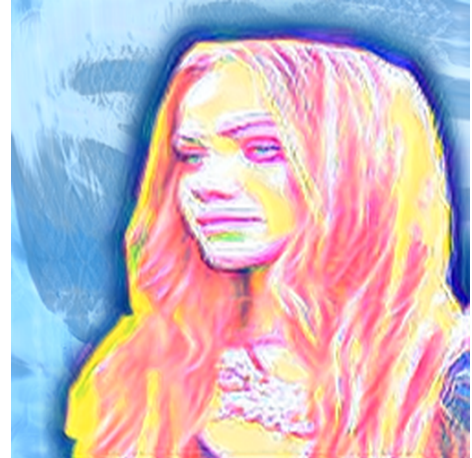
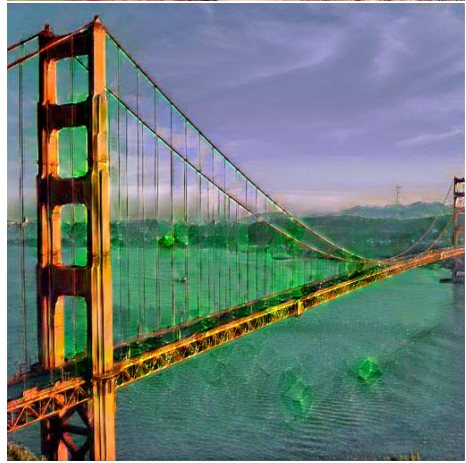
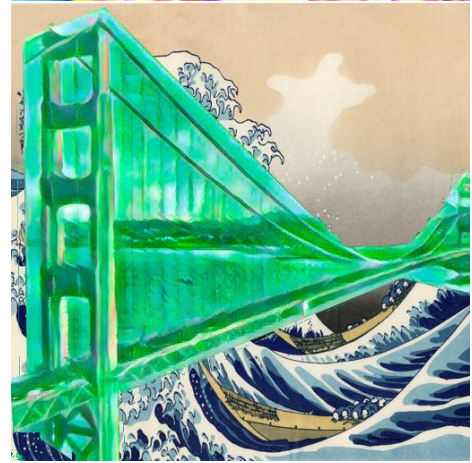

    \centering
      \begin{minipage}{0.15\linewidth}         
    \centering
        \textbf{F:} Fantasy Vivid Colours. \\
        \textbf{B:} Starry Night.
        \end{minipage}         
    \begin{minipage}{0.35\linewidth}
         \centering             \includegraphics[width=0.99\linewidth]{images/SemCSglobfb_Fantasyvividcolors_StarryNight_0976_resized.jpg}
        \end{minipage}
    \begin{minipage}{0.35\linewidth}
         \centering             \includegraphics[width=0.99\linewidth]{images/Fantasyvividcolors_0976_resized_ours.jpg}
        \end{minipage}
    \begin{minipage}{0.15\linewidth}         
    \centering
        \textbf{F:} Lisa Frank. \\
        \textbf{B:} Ice.
    \end{minipage}           
    \begin{minipage}{0.35\linewidth}
         \centering             \includegraphics[width=0.99\linewidth]{images/SemCSglobfb_Ice_LisaFrank_blonde_girl.jpg}
        \end{minipage}
    \begin{minipage}{0.35\linewidth}
         \centering             \includegraphics[width=0.99\linewidth]{images/Ice_LisaFrank_blonde_girl.png}
        \end{minipage}
       \begin{minipage}{0.15\linewidth}         
    \centering
        \textbf{F:} Green Crystal \\
        \textbf{B:} The great wave off King Kanagawa.
    \end{minipage}           
    \begin{minipage}{0.35\linewidth}
         \centering             \includegraphics[width=0.99\linewidth]{images/SemCSglobfb_TheGreatWaveof_GreenCrystal_golden_gate.jpg}
        \end{minipage}
    \begin{minipage}{0.35\linewidth}
         \centering             \includegraphics[width=0.99\linewidth]{images/GreenCrystal_golden_gate_ours.jpg}
        \end{minipage}
        \caption{\textbf{TIST (Double)}}
        \label{fig:fig4}
\end{figure*}

\noindent \textbf{Text-Guided IST Outputs (Double).} The outputs presented in Fig. \ref{fig:fig3} and \ref{fig:fig4} are the extended view of Figure 6 of the main manuscript. We compared our outputs with SemCS \cite{kamra2023sem}. SemCS fails to apply distinct style features on salient objects and surrounding elements very well. There is no distinction of style feature boundaries in the foreground and background of the content image. Hence our method presents outputs superior to SemCS. \\~

\begin{figure*}
    \centering
       \begin{minipage}{0.20\linewidth}
       \centering
            Style Text 
        \end{minipage}
    \begin{minipage}{0.19\linewidth}
    \centering
            Style Image
        \end{minipage}
    \begin{minipage}{0.23\linewidth}
    \centering
            MMIST \cite{Wang2024WACV}
        \end{minipage}
    \begin{minipage}{0.23\linewidth}
    \centering
            Ours
        \end{minipage}      
    \begin{minipage}{0.20\linewidth}
         Copper Plate Engraving.   
        \end{minipage} 
    \begin{minipage}{0.19\linewidth}
         \centering             \includegraphics[width=0.99\linewidth]{images/contrast_of_forms.png}
        \end{minipage}
        \begin{minipage}{0.23\linewidth}
         \centering             \includegraphics[width=0.99\linewidth]{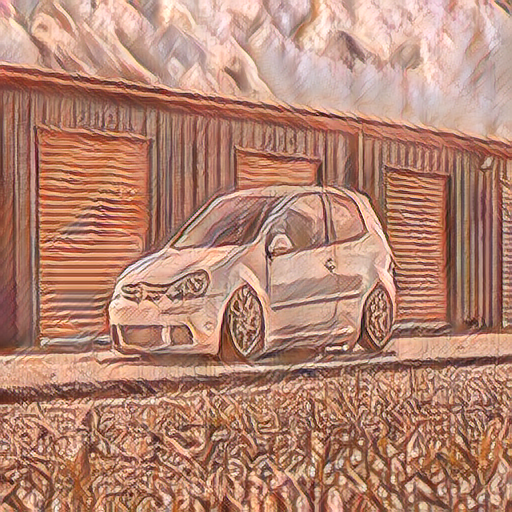}
        \end{minipage}
        \begin{minipage}{0.23\linewidth}
         \centering             \includegraphics[width=0.99\linewidth]{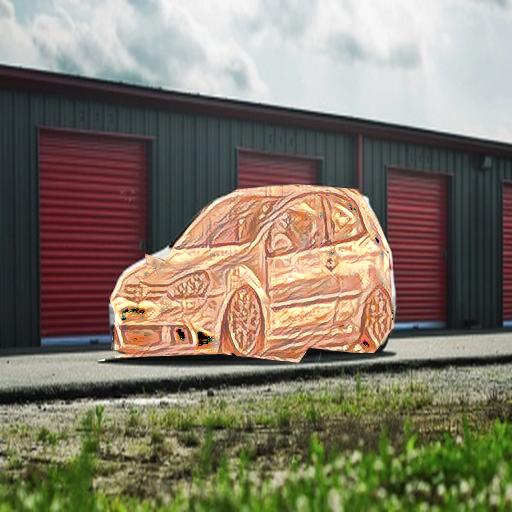}
        \end{minipage}
    % \begin{minipage}{0.30\linewidth}
    % \begin{overpic}[width=\textwidth]{images/multimodal/picasso_seated_nude_hr.png}
    % % \put(10,90){\textcolor{white}{\Huge Your Text Here}}
    % \put(10,30){\textcolor{white}{\scriptsize The Great Wave off Kanagawa by Katsushika Hokusai.}}
    % % Adjust the coordinates (10, 90) and (50, 50) as needed
    % \end{overpic}
    % \end{minipage}
      \begin{minipage}{0.20\linewidth}
         \centering
         % \scriptsize
        The Great Wave off Kanagawa by Katsushika Hokusai.       
    \end{minipage}       
     \begin{minipage}{0.19\linewidth}
         \centering
             \includegraphics[width=0.99\linewidth]{images/picasso_seated_nude_hr.png}
        \end{minipage}
    \begin{minipage}{0.23\linewidth}
         \centering
             \includegraphics[width=0.99\linewidth]{images/ThegreatwaveoffKanagawaHoksuai_0958_resized_mmist.jpg}
        \end{minipage}
 \begin{minipage}{0.23\linewidth}
         \centering
             \includegraphics[width=0.99\linewidth]{images/ThegreatwaveoffKanagawaHoksuai_0958_resized_ours.jpg}
        \end{minipage}       
     \begin{minipage}{0.20\linewidth}
         \centering
         % \small
        Impressionism        
    \end{minipage}
    \begin{minipage}{0.19\linewidth}
         \centering
             \includegraphics[width=0.99\linewidth]{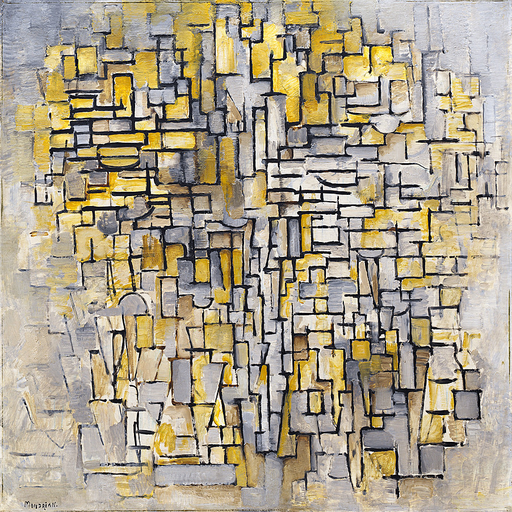}
        \end{minipage} 
        \begin{minipage}{0.23\linewidth}
         \centering
             \includegraphics[width=0.99\linewidth]{images/Impressionism_0958_resized_mmist.jpg}
        \end{minipage}
        \begin{minipage}{0.23\linewidth}
         \centering
             \includegraphics[width=0.99\linewidth]{images/Impressionism_0958_resized_ours.jpg}
        \end{minipage}
     \begin{minipage}{0.20\linewidth}
         \centering
         % \small
        A Baroque Painting        
    \end{minipage}
      \begin{minipage}{0.19\linewidth}
         \centering
             \includegraphics[width=0.99\linewidth]{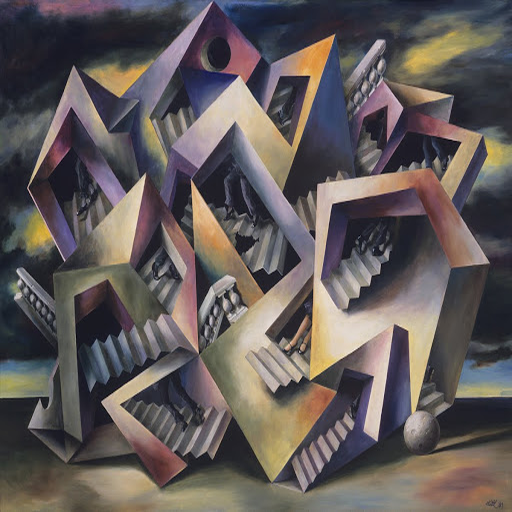}
        \end{minipage}
    \begin{minipage}{0.23\linewidth}
         \centering
             \includegraphics[width=0.99\linewidth]{images/ABaroquepainting_0958_resized_mmist.jpg}
        \end{minipage}
 \begin{minipage}{0.23\linewidth}
         \centering
             \includegraphics[width=0.99\linewidth]{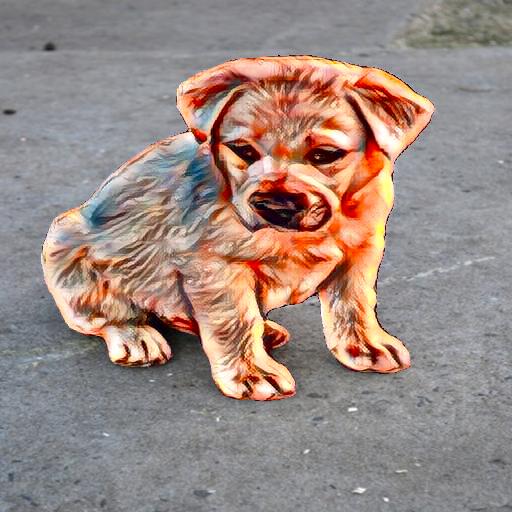}
        \end{minipage}
     \begin{minipage}{0.20\linewidth}
         \centering
         \small
        Lisa Frank
    \end{minipage}
    \begin{minipage}{0.19\linewidth}
         \centering
             \includegraphics[width=0.99\linewidth]{images/antimonocromatismo.png}
        \end{minipage}
    \begin{minipage}{0.23\linewidth}
         \centering
             \includegraphics[width=0.99\linewidth]{images/LisaFrank_0958_resized_mmist.jpg}
    \end{minipage}
        \begin{minipage}{0.23\linewidth}
         \centering
             \includegraphics[width=0.99\linewidth]{images/LisaFrank_0958_resized_ours.jpg}
        \end{minipage}      
    \caption{\textbf{Multimodal IST:} Column 1 and 2 are multimodal inputs for style features. Column 3 and 4 are stylized outputs.}
    \label{fig:fig5}
\end{figure*}

\noindent \textbf{Multimodality-Guided IST Outputs (Single).} In Fig. \ref{fig:fig5}, we perform stylization with multimodal inputs (style text and image) and compare our results with recent baseline method MMIST \cite{Wang2024WACV}. Our method performs object-specific stylization only on flower, whereas MMIST applies the style features entirely on the image (flower and background) causing to content mismatch problem. We solve this problem with image harmonization. 

% \vspace{-1.0cm}
\section{Generating Style Representations}
% \vspace{-0.2cm}
\noindent \textbf{Drawback of All-to-All Attention Mapping.}
With an all-to-all attention mechanism, dense correspondence gets established between content and style features during style transfer. A small change in the position of elements significantly affects the attention results. This happens because the softmax function used in attention calculations involves exponential computations, it tends to focus strongly on the most similar value, leading to high exclusivity in attention scores. When the most similar key (the target or focus of attention) is semantically different from the query (an element in the input data), it results in distorted style patterns. This signifies that the all-to-all attention mechanism is highly sensitive to the positions of elements within the input data. \\~
\noindent \textbf{Salient-To-Key (S2K).} To overcome, these issues, we propose to use Salient-to-Key mapping, which leverages a sparse connection between content and style features through distributive and progressive attention as follows:
\begin{enumerate}
    \item \textit{Distributive attention (DA).} With DA, the model learns distributed keys, which capture and represent the style distribution of various local regions within the style features. Instead of relying on individual, it uses these distributed keys to create a more robust representation of style.  Each query from the content features (the input data that needs to be stylized) is matched with these stable and representative distributed keys. This matching process leverages the learned distribution to ensure a more stable and accurate style transfer. DA enhances the model’s tolerance to matching errors. This means that even if there is some variability or noise in the input, the model can still perform accurate style transfer because it focuses on regional patterns. 
    \item \textit{Progressive Attention (PA).} With this setting, the attention mechanism starts by focusing on broader, coarse-grained regions of the input and progressively narrows down to fine-grained details. This hierarchical approach ensures that the model captures both the overall structure and the intricate details of the input data. Queries within a local region of the content features are matched to the same stable keys within that local region. This consistency helps in maintaining the local semantic integrity during style transfer. 
    PA ensures that the attention mechanism starts with a broad overview and then refines its focus. 
\end{enumerate}
\noindent \textbf{Cross Modal GAN Inversion.} Our framework ObjMST proposes style-specific "masked" directional CLIP loss $L_{fg}$ to generate style representations $S_{fg}$. This objective function identifies those aligned directions in the CLIP-embedding space that disentangle content features on input style-image and intermediate style representations. To obtain the S, latent vector $w_{fg}$ is initialized randomly and is passed to StyleGAN3 $G$, where input style image ($I_S$) and intermediate style representation S gets cropped and augmented in cross-modal GAN inversion as shown in Fig.~\ref{fig:crossmodal}.
\begin{equation}
    S_{fg}=G(w_{fg})    
\end{equation}
\begin{equation}
    \left[\{S^j\}, \{I_{S_{fg}}^k\}\}\right]_{j,k=1}^{N_{crop}} = aug(crop(S,I_S))
\end{equation}
$aug(.)$ and $crop(.)$ are the augmentation and crop function respectively; and $N_{crop}$ is the number of cropped patches.  The rest of the details for computing style-specific masked directional CLIP loss are present in the main manuscript.  
% TIST (Single)

% TIST (Double)

% MMIST (Double)
% \vspace{20.2cm}
% NIMA

\begin{figure*}[!htb]
\begin{minipage}{0.49\textwidth}
\scriptsize
\vspace{0.7cm}
\captionof{table}{NIMA Score}
\vspace{0.3cm}
\begin{tabular}{|c|c|c|c|c|c|}
\hline
\rowcolor[HTML]{FFFFFF} 
\cellcolor[HTML]{FFFFFF}{\color[HTML]{333333} } & {\color[HTML]{333333} \textbf{CS} \cite{kwon2022clipstyler}} & \textbf{SemCS} \cite{kamra2023sem} & \textbf{LDAST} \cite{fu2022language} & \textbf{MMIST} \cite{Wang2024WACV} & \textbf{Ours} \\ \hline
1                                                             & 6.28                               & 3.16           & 0.44           & 0.73           & 8.98          \\ \hline
2                                                             & 9.71                               & 2.89           & 4.24           & 5.64           & 8.62          \\ \hline
3                                                             & 7.05                               & 9.27           & 5.76           & 6.57           & 10.25         \\ \hline
4                                                             & 7.56                               & 1.08           & 1.49           & 6.65           & 1.00          \\ \hline
5                                                             & 0.85                               & 2.45           & 7.85           & 5.56           & 8.22          \\ \hline
6                                                             & 6.10                               & 2.57           & 8.97           & 2.59           & 2.57          \\ \hline
7                                                             & 6.75                               & 6.00           & 7.51           & 8.30           & 6.56          \\
\hline
\end{tabular}
\end{minipage}%
\begin{minipage}{0.49\textwidth}
\scriptsize
% \hspace{24cm}
% \vspace{4cm}
\begin{tabular}{|c|c|c|c|c|c|}
\hline
\rowcolor[HTML]{FFFFFF} 
\cellcolor[HTML]{FFFFFF}{\color[HTML]{333333} } & {\color[HTML]{333333} \textbf{CS} \cite{kwon2022clipstyler}} & \textbf{SemCS} \cite{kamra2023sem} & \textbf{LDAST} \cite{fu2022language} & \textbf{MMIST} \cite{Wang2024WACV} & \textbf{Ours} \\ \hline
65                                                            & 3.10                               & 0.68           & 5.85           & 5.84           & 3.28          \\ \hline
66                                                            & 8.28                               & 6.74           & 7.16           & 2.61           & 9.29          \\ \hline
67                                                            & 9.60                               & 5.03           & 1.20           & 4.50           & 6.89          \\ \hline
68                                                            & 1.08                               & 2.51           & 1.86           & 3.49           & 4.51          \\ \hline
69                                                            & 1.32                               & 4.72           & 3.33           & 6.11           & 4.31          \\ \hline
70                                                            & 5.77                               & 3.15           & 4.86           & 3.90           & 5.87          \\ \hline
71                                                            & 9.76                               & 2.20           & 9.75           & 1.86           & 0.47          \\
\hline
\end{tabular}
\end{minipage}
\end{figure*}

% ContriqueFR
\begin{figure*}[!htb]
\begin{minipage}{0.49\textwidth}
\scriptsize
\vspace{0.7cm}
\captionof{table}{Contrique-FR Score}
\vspace{0.3cm}
\begin{tabular}{|c|c|c|c|c|c|}
\hline
\rowcolor[HTML]{FFFFFF} 
\cellcolor[HTML]{FFFFFF}{\color[HTML]{333333} } & {\color[HTML]{333333} \textbf{CS} \cite{kwon2022clipstyler}} & \textbf{SemCS} \cite{kamra2023sem} & \textbf{LDAST} \cite{fu2022language} & \textbf{MMIST} \cite{Wang2024WACV} & \textbf{Ours} \\ \hline
1                                                             & 0.46                               & 0.30           & 0.58           & 0.49           & 0.60          \\ \hline
2                                                             & 0.21                               & 0.34           & 0.07           & 0.32           & 0.64          \\ \hline
3                                                             & 0.48                               & 0.09           & 0.56           & 0.04           & 0.04          \\ \hline
4                                                             & 0.44                               & 0.18           & 0.51           & 0.36           & 0.45          \\ \hline
5                                                             & 0.37                               & 0.44           & 0.58           & 0.23           & 0.32          \\ \hline
6                                                             & 0.30                               & 0.34           & 0.22           & 0.56           & 0.22          \\ \hline
7                                                             & 0.30                               & 0.34           & 0.28           & 0.24           & 0.24          \\
\hline
\end{tabular}
\end{minipage}%
\begin{minipage}{0.49\textwidth}
\scriptsize
\vspace{4cm}
\begin{tabular}{|c|c|c|c|c|c|}
\hline
\rowcolor[HTML]{FFFFFF} 
\cellcolor[HTML]{FFFFFF}{\color[HTML]{333333} } & {\color[HTML]{333333} \textbf{CS} \cite{kwon2022clipstyler}} & \textbf{SemCS} \cite{kamra2023sem} & \textbf{LDAST} \cite{fu2022language} & \textbf{MMIST} \cite{Wang2024WACV} & \textbf{Ours} \\ \hline
65                                                            & 0.21                               & 0.04           & 0.37           & 0.37           & 0.19          \\ \hline
66                                                            & 0.32                               & 0.32           & 0.15           & 0.40           & 0.37          \\ \hline
67                                                            & 0.16                               & 0.03           & 0.24           & 0.22           & 0.07          \\ \hline
68                                                            & 0.26                               & 0.02           & 0.02           & 0.48           & 0.24          \\ \hline
69                                                            & 0.38                               & 0.12           & 0.42           & 0.27           & 0.58          \\ \hline
70                                                            & 0.47                               & 0.16           & 0.05           & 0.38           & 0.07          \\ \hline
71                                                            & 0.48                               & 0.28           & 0.52           & 0.15           & 0.22          \\
\hline
\end{tabular}
\end{minipage}
\end{figure*}

\begin{figure*}[!htb]
% Contrique NFR
\begin{minipage}{0.49\textwidth}
\scriptsize
\vspace{0.7cm}
\captionof{table}{Contrique NFR}
\vspace{0.3cm}
\begin{tabular}{|c|c|c|c|c|c|}
\hline
\rowcolor[HTML]{FFFFFF} 
\cellcolor[HTML]{FFFFFF}{\color[HTML]{333333} } & {\color[HTML]{333333} \textbf{CS} \cite{kwon2022clipstyler}} & \textbf{SemCS} \cite{kamra2023sem} & \textbf{LDAST} \cite{fu2022language} & \textbf{MMIST} \cite{Wang2024WACV} & \textbf{Ours} \\ \hline
1                                                             & 14.06                              & 33.84          & 78.45          & 81.26          & 26.44         \\ \hline
2                                                             & 28.08                              & 50.12          & 34.80          & 8.13           & 84.86         \\ \hline
3                                                             & 27.25                              & 21.50          & 29.70          & 71.73          & 3.21          \\ \hline
4                                                             & 14.77                              & 13.90          & 22.91          & 5.20           & 82.98         \\ \hline
5                                                             & 2.11                               & 46.66          & 90.21          & 35.89          & 22.81         \\ \hline
6                                                             & 31.86                              & 25.64          & 73.58          & 88.24          & 38.18         \\ \hline
7                                                             & 10.37                              & 37.24          & 45.99          & 45.92          & 67.62         \\
\hline
\end{tabular}
\end{minipage}
\begin{minipage}{0.49\textwidth}
% \caption{Contrique-NFR}
\scriptsize
\vspace{4cm}
\begin{tabular}{|c|c|c|c|c|c|}
\hline
\rowcolor[HTML]{FFFFFF} 
\cellcolor[HTML]{FFFFFF}{\color[HTML]{333333} } & {\color[HTML]{333333} \textbf{CS} \cite{kwon2022clipstyler}} & \textbf{SemCS} \cite{kamra2023sem} & \textbf{LDAST} \cite{fu2022language} & \textbf{MMIST} \cite{Wang2024WACV} & \textbf{Ours}  \\ \hline
65                                                            & 41.57                              & 12.34          & 68.51          & 77.48          & 15.70          \\ \hline
66                                                            & 16.83                              & 12.80          & 10.32          & 19.75          & 33.06          \\ \hline
67                                                            & 49.91                              & 5.64           & 14.80          & 9.91           & 72.83          \\ \hline
68                                                            & 31.09                              & 8.54           & 43.72          & 69.36          & 30.04          \\ \hline
69                                                            & 21.82                              & 48.30          & 59.12          & 61.37          & 77.17          \\ \hline
70                                                            & 59.92                              & 7.72           & 42.82          & 55.74          & 87.04          \\ \hline
71                                                            & 26.79                              & 52.31          & 68.44          & 83.82          & 50.44          \\
\hline
\end{tabular}
\end{minipage}
\end{figure*}

\end{document}